\pdfoutput=1

\documentclass[11pt]{article}

\usepackage{acl}

\usepackage{times}
\usepackage{latexsym}

\usepackage[T1]{fontenc}

\usepackage[utf8]{inputenc}

\usepackage{microtype}

\usepackage{amssymb}
\usepackage{amsmath}
\usepackage{multirow,booktabs, hhline}
\usepackage{dsfont}
\usepackage{wrapfig}
\usepackage{subfigure}

\usepackage{epigraph}
\epigraphsize{\small}
\usepackage{graphicx}
\usepackage{cleveref}
\crefname{section}{§}{§§}
\Crefname{section}{§}{§§}

\title{Knowledge Inheritance for Pre-trained Language Models}
\author{
 Yujia~Qin$^{1,2,3}$, Yankai~Lin$^{4}$, Jing~Yi$^{1,2,3}$, Jiajie~Zhang$^{1,2,3}$, \textbf{Xu~Han}$^{1,2,3}$, \\ \textbf{Zhengyan~Zhang}$^{1,2,3}$, \textbf{Yusheng~Su}$^{1,2,3}$, \textbf{Zhiyuan~Liu}$^{1,2,3,5,6}$\thanks{\ \  Corresponding author.}, \textbf{Peng~Li}$^7$\thanks{\ \ Part of the work was done while Peng Li was working at Tencent.}, \\ \textbf{Maosong~Sun$^{1,2,3,5*}$, Jie~Zhou$^4$} \\
 $^1$Department of Computer Science and Technology, Tsinghua University, Beijing, China \\
 $^2$Beijing National Research Center for Information Science and Technology \\
 $^3$Institute for Artificial Intelligence, Tsinghua University, Beijing, China\\
 $^4$Pattern Recognition Center, WeChat AI, Tencent Inc. \\
 $^5$International Innovation Center of Tsinghua University, Shanghai, China \\
 $^6$Quan Cheng Laboratory \\
 $^7$Institute for AI Industry Research (AIR), Tsinghua University, China. \\
\texttt{qyj20@mails.tsinghua.edu.cn}\\
}

\begin{document}

\maketitle
\begin{abstract}
Recent explorations of large-scale pre-trained language models (PLMs) have revealed the power of PLMs with huge amounts of parameters, setting off a wave of training ever-larger PLMs. However, it requires tremendous computational resources to train a large-scale PLM, which may be practically unaffordable. In addition, existing large-scale PLMs are mainly trained from scratch individually, ignoring that many well-trained PLMs are available. To this end, we explore the question how could existing PLMs benefit training large-scale PLMs in future. Specifically, we introduce a pre-training framework named ``knowledge inheritance'' (KI) and explore how could knowledge distillation serve as auxiliary supervision during pre-training to efficiently learn larger PLMs. Experimental results demonstrate the superiority of KI in training efficiency. We also conduct empirical analyses to explore the effects of teacher PLMs' pre-training settings, including model architecture, pre-training data, etc. Finally, we show that KI could be applied to domain adaptation and knowledge transfer. The implementation is publicly available at \url{https://github.com/thunlp/Knowledge-Inheritance}.
\end{abstract}
\section{Introduction}
Recently, it has become a consensus in the NLP community to use pre-trained language models (PLMs) as the backbone for various downstream tasks~\citep{Han2021PreTrainedMP,min2021recent}. Despite the great follow-up efforts of exploring various pre-training techniques and model architectures, researchers find that simply enlarging the model capacity, data size and training steps can further improve the performance of PLMs~\cite{kaplan2020scaling,li2020train}. This discovery sets off a wave of training large-scale PLMs~\cite{raffel2019exploring,NEURIPS2020_1457c0d6,fedus2021switch}. 


Although huge PLMs have shown awesome performance~\cite{bommasani2021opportunities}, it requires tremendous computational resources to train large-scale PLMs~\cite{schwartz2019green}, raising severe environmental concerns on the prohibitive computational costs. Moreover, existing PLMs are generally trained from scratch individually, ignoring that many well-trained PLMs are available. This leaves us an important question: \textit{how could existing PLMs benefit training larger PLMs in future?}

Considering that humans can leverage the knowledge summarized by their predecessors to learn new tasks, so that the learning process could become efficient; similarly, it is worth inheriting the implicit knowledge distributed in existing PLMs. In this sense, we could distill the knowledge summarized by an existing small PLM during pre-training to efficiently learn larger PLMs. We dub the above process as \textit{knowledge inheritance} (KI). This intuition is similar to reversed KD~\citep{yuan2020revisiting} in the field of computer vision. They indicate that a delicate student model could still benefit from a teacher with an inferior architecture for a specific downstream task. 

However, the success of reversed KD in supervised downstream tasks does not guarantee its feasibility under the scenario of large-scale self-supervised pre-training. 
Therefore, in this paper, we strive to answer the following research questions: (\textit{RQ1}) could distilling knowledge from an existing trained PLM benefit large PLMs' training from scratch? (\textit{RQ2}) Considering human beings are able to hand down knowledge from generation to generation, could KI similarly be sequentially performed among a series of PLMs with growing sizes? (\textit{RQ3}) As more and more PLMs with different pre-training settings (model architectures, training data, training strategies, etc) emerge, how would different settings affect the performance of KI? (\textit{RQ4}) Besides training a large PLM from scratch, when adapting an already trained large PLM to a new domain, how could smaller domain teachers benefit such a process?

In conclusion, the contributions of this paper are summarized as follows: (1) we are the first to formulate the problem of knowledge inheritance, and demonstrate the feasibility of inheriting the knowledge from previously trained PLMs for efficiently training larger ones; (2) we show that the learned knowledge in PLMs could accumulate and further be passed down from generation to generation; (3) we systematically conduct empirical analyses to show the effects of various teacher pre-training settings, which may indicate how to select the most appropriate PLM as the teacher for KI; (4) we further show that during domain adaptation, an already trained large PLM could benefit from multiple small PLMs of different domains under the KI framework. The above empirical studies indicate that KI can well support cross-model knowledge transfer, providing a promising direction to share the knowledge learned by different PLMs and continuously promote their performance.

\section{Related Work}

\paragraph{Efficient Pre-training for NLP.} Recently, researchers find that the performance of PLMs can be simply improved by increasing the model size, data size and training steps~\cite{liu2019roberta,raffel2019exploring,kaplan2020scaling}, sparking a wave of training ever-larger PLMs. For instance, the revolutionary GPT-3~\cite{NEURIPS2020_1457c0d6}, which contains $175$ billion parameters, shows strong capabilities for language understanding and generation. This means that utilizing PLMs with huge parameters for downstream tasks may greatly relieve the cost of manual labeling and model training for new tasks. However, larger models require greater computational demands~\cite{patterson2021carbon}. To this end, researchers propose to accelerate pre-training by mixed-precision training~\cite{shoeybi2019megatron}, distributed training~\cite{shoeybi2019megatron}, large batch optimization~\cite{you2019large}, etc.

Another line of methods~\cite{gong2019efficient,gu2020transformer,chen2021bert2bert,qin2022elle} proposes to pre-train larger PLMs progressively. They first train a small PLM, and then gradually increase the depth or width of the network based on parameter recycling (PR). Although PR could be used for the goal of KI, these methods typically have strict requirements on the architectures of both models, which is not flexible for practical uses; instead, we resort to KD as the solution for KI without architecture constraints. In addition, different from KI, PR is not applicable for absorbing knowledge from multiple teacher models and domain adaptation. More detailed comparisons between KI and PR are discussed in \cref{sec:compare_KI_PT}.

\paragraph{Knowledge Distillation for PLMs.} Knowledge Distillation (KD)~\cite{hinton2015distilling} aims to compress a large model into a fast-to-execute one. KD has renewed a surge of interest in PLMs recently. Some explore KD at different training phases, e.g., pre-training~\cite{sanh2019distilbert}, downstream fine-tuning~\cite{sun2019patient,krishna2019thieves}, or both of them~\cite{jiao2019tinybert}; others explore distilling not only the final logits output by the large PLM, but also the intermediate hidden representations~\cite{sanh2019distilbert,jiao2019tinybert,sun2020contrastive}. Conventional KD presumes that teacher models play pivotal roles in mastering knowledge, and student models generally cannot match their teachers in performance. When it comes to the scenario of KI, since student models have larger capacities, the performance of teacher models is no longer an ``upper bound'' of student models. Outside NLP, researchers recently demonstrate that a student model could also benefit from a poor teacher for a specific downstream task~\citep{yuan2020revisiting} (reversed KD). Based on the prior explorations, in this paper, we investigate the application of reversed KD in pre-training. 
\section{Knowledge Inheritance}
\paragraph{Task Formulation.} Given a textual input $\mathbf{x} = \{x^1,\ldots,x^n\}$ and the corresponding label $\mathbf{y} \in \mathbb{R}^K$, where $K$ is the number of classes for the specific pre-training task, e.g., the vocabulary size for masked language modeling (MLM)~\cite{devlin2018bert}, a PLM $\mathcal{M}$ converts each token $x^j \in \mathbf{x}$ to task-specific logits $\mathbf{z}^j = [z_1^j, ..., z_K^j]$. $\mathbf{z}^j$ is then converted to a probability distribution $\mathcal{P}(x^j; \tau) = [p_1(x^j; \tau), ..., p_K(x^j; \tau)]$ using a softmax function with temperature $\tau$. $\mathcal{M}$ is pre-trained with the objective $\mathcal{L}_{\text{SELF}}(\mathbf{x}, \mathbf{y}) = \mathcal{H}(\mathbf{y}, \mathcal{P}(\mathbf{x}; \tau))$, where $\mathcal{H}$ is the loss function, e.g., cross-entropy for MLM. Assume that we have a well-trained small PLM $\mathcal{M}_S$ optimized with the self learning objective $\mathcal{L}_{\text{SELF}}$ (such as MLM), our goal is leveraging $\mathcal{M}_S$'s knowledge to efficiently train a larger PLM $\mathcal{M}_L$ on the corpora $\mathcal{D}_L = \{(\mathbf{x}_i, \mathbf{y}_i)\}_{i=1}^{|\mathcal{D}_L|}$.


\paragraph{Investigated Methodology.} 
Specifically, imparting $\mathcal{M}_S$'s knowledge to $\mathcal{M}_L$ on $\mathcal{D}_L$ is implemented by minimizing the Kullback-Leibler (KL) divergence between two probability distributions output by $\mathcal{M}_S$ and $\mathcal{M}_L$ on the same input $\mathbf{x}_i \in \mathcal{D}_L$, i.e., $\mathcal{L}_{\text{KI}}(\mathbf{x}_i; \mathcal{M}_S) = \tau^2\text{KL}(\mathcal{P}_{\mathcal{M}_S}(\mathbf{x}_i; \tau) || \mathcal{P}_{\mathcal{M}_L}(\mathbf{x}_i; \tau)))$. In addition, $\mathcal{M}_L$ is also encouraged to conduct self-learning by optimizing $\mathcal{L}_{\text{SELF}}(\mathbf{x}_i, \mathbf{y}_i)$. Both $\mathcal{L}_{\text{SELF}}$ and $\mathcal{L}_{\text{KI}}$ are balanced with an inheritance rate $\alpha$:
\begin{equation*}
\small
\begin{aligned}
  \mathcal{L}(\mathcal{D}_L; \mathcal{M}_S)
    &= \!\!\!\!\!\! \sum_{(\textbf{x}_i, \textbf{y}_i) \in \mathcal{D}_L} \!\!\!\!\!\! (1 - \alpha)  \mathcal{L}_{\text{SELF}}(\textbf{x}_i, \textbf{y}_i)
  + \alpha  \mathcal{L}_{\text{KI}}(\textbf{x}_i; \mathcal{M}_S) \\
\end{aligned}
\end{equation*}
\vspace{-1.55em}
\begin{equation}
\small
\begin{aligned}
  \quad \quad \quad \! \! & = \!\!\!\!\!\! \sum_{(\textbf{x}_i, \textbf{y}_i) \in \mathcal{D}_L} \!\!\!\!\!\! (1 - \alpha) \mathcal{H}(\mathbf{y}_i, \mathcal{P}_{\mathcal{M}_L}(\mathbf{x}_i; 1)) \\
  & + \alpha \tau^2  \text{KL}(\mathcal{P}_{\mathcal{M}_S}(\mathbf{x}_i; \tau) || \mathcal{P}_{\mathcal{M}_L}(\mathbf{x}_i; \tau))).
\end{aligned}
\end{equation}

Since larger models generally converge faster and can achieve better final performance~\cite{li2020train}, $\mathcal{M}_L$ becomes more and more knowledgeable during the learning process, and would surpass the teacher eventually. Thus, it is necessary to encourage $\mathcal{M}_L$ increasingly learning knowledge on its own, not only following the teacher's instructions. Additionally, after $\mathcal{M}_L$ has surpassed its teacher, it no longer needs the guidance from $\mathcal{M}_S$ and should conduct pure self-learning from then on. Therefore, different from reversed KD, we dynamically change the inheritance rate $\alpha$. Specifically, for a total training steps of $T$, we linearly decay $\alpha_t$ with a slope of $\frac{\alpha_T}{T}$. The student only inherits knowledge from the teacher for $\frac{T}{\alpha_T}$ steps, and then conducts pure self-learning, i.e., $\alpha_t = \max(1 - \alpha_T \times \frac{t}{T}, 0)$. Formally, at step $t$, the loss function for inheriting knowledge of $\mathcal{M}_S$ on $\mathcal{D}_L$ is formulated as:
\begin{equation}
\label{KI_eq}
\small
\begin{aligned}
  \mathcal{L}(\mathcal{D}_L; \mathcal{M}_S) =  \!\!\!\!\!\!\!\! \sum_{(\textbf{x}_i, \textbf{y}_i) \in \mathcal{D}_L} \!\!\!\!\!\! (1 \!-\! \alpha_t)  \mathcal{L}_{\text{SELF}}(\textbf{x}_i, \textbf{y}_i) \!+\! \alpha_t  \mathcal{L}_{\text{KI}}(\textbf{x}_i; \mathcal{M}_S).
\end{aligned}
\end{equation}


Note the logits of $\mathcal{M}_S$ on $\mathcal{D}_L$ can be pre-computed and saved offline so that we do not need to re-compute the logits of $\mathcal{M}_S$ when training $\mathcal{M}_L$. This process is done once and for all. 

\section{Empirical Analysis}
In this section, we answer our research questions proposed before. Specifically, (1) we first demonstrate the effectiveness of KI in \cref{sec:prelim}. (2) Then we show PLMs can accumulate knowledge over generations in \cref{generation}. (3) We also investigate the effects of different pre-training settings of the teacher models in \cref{sec:effects}. (4) Finally, we show that KI could benefit domain adaptation, and a trained PLM can learn more efficiently with the help of multiple domain teachers in \cref{sec:continual}. Detailed pre-training hyper-parameters are listed in \cref{sec:hyper_pretrain}. 

\begin{figure*}[!t]
\vfill
\centerline{
\begin{minipage}{0.3\textwidth}
    \centerline{\includegraphics[width=\textwidth]{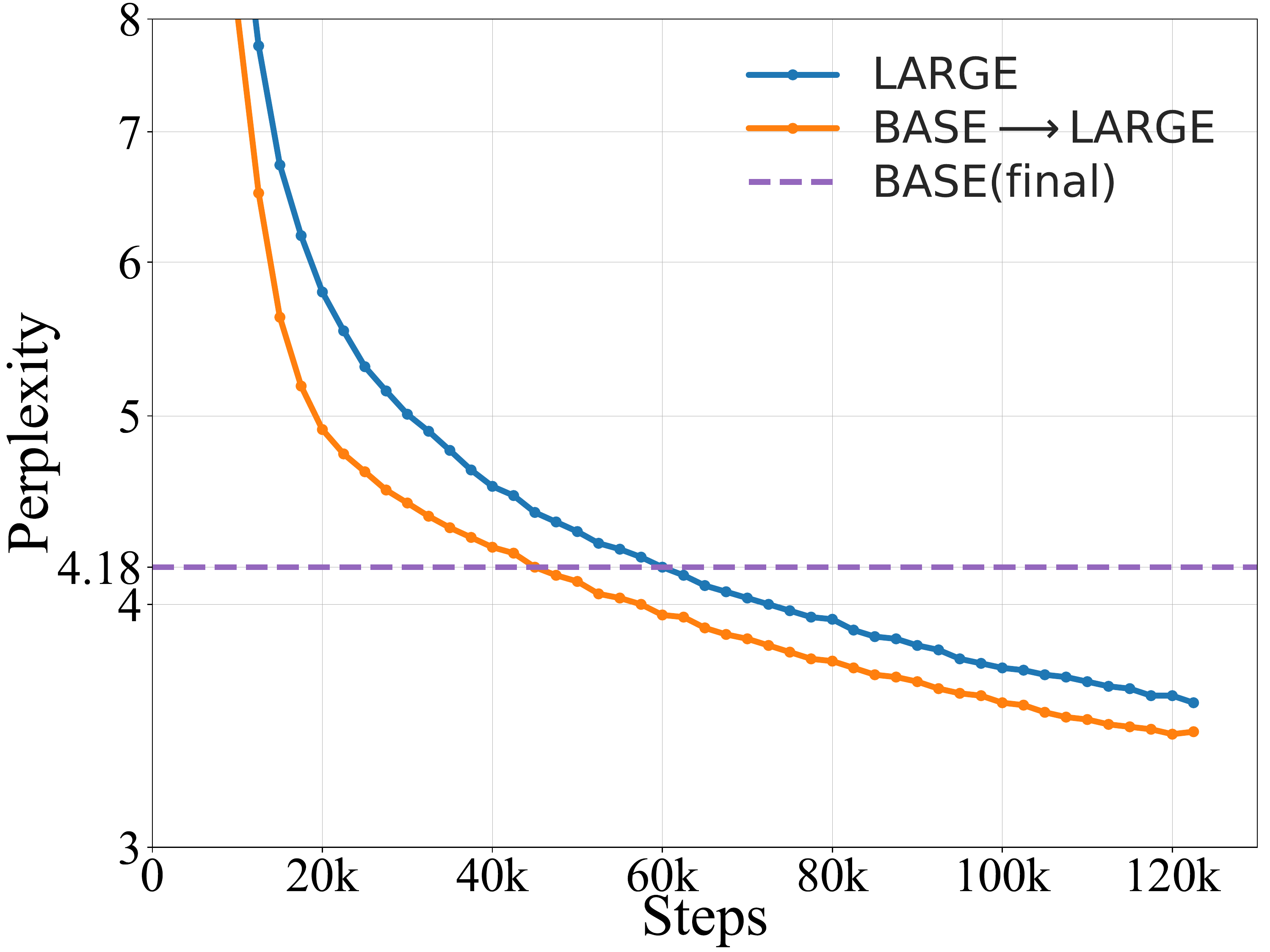}}
\centerline{\footnotesize{(a)}}
\end{minipage}
\hspace{0pt}
\begin{minipage}{0.3\textwidth}
\centerline{\includegraphics[width=\textwidth]{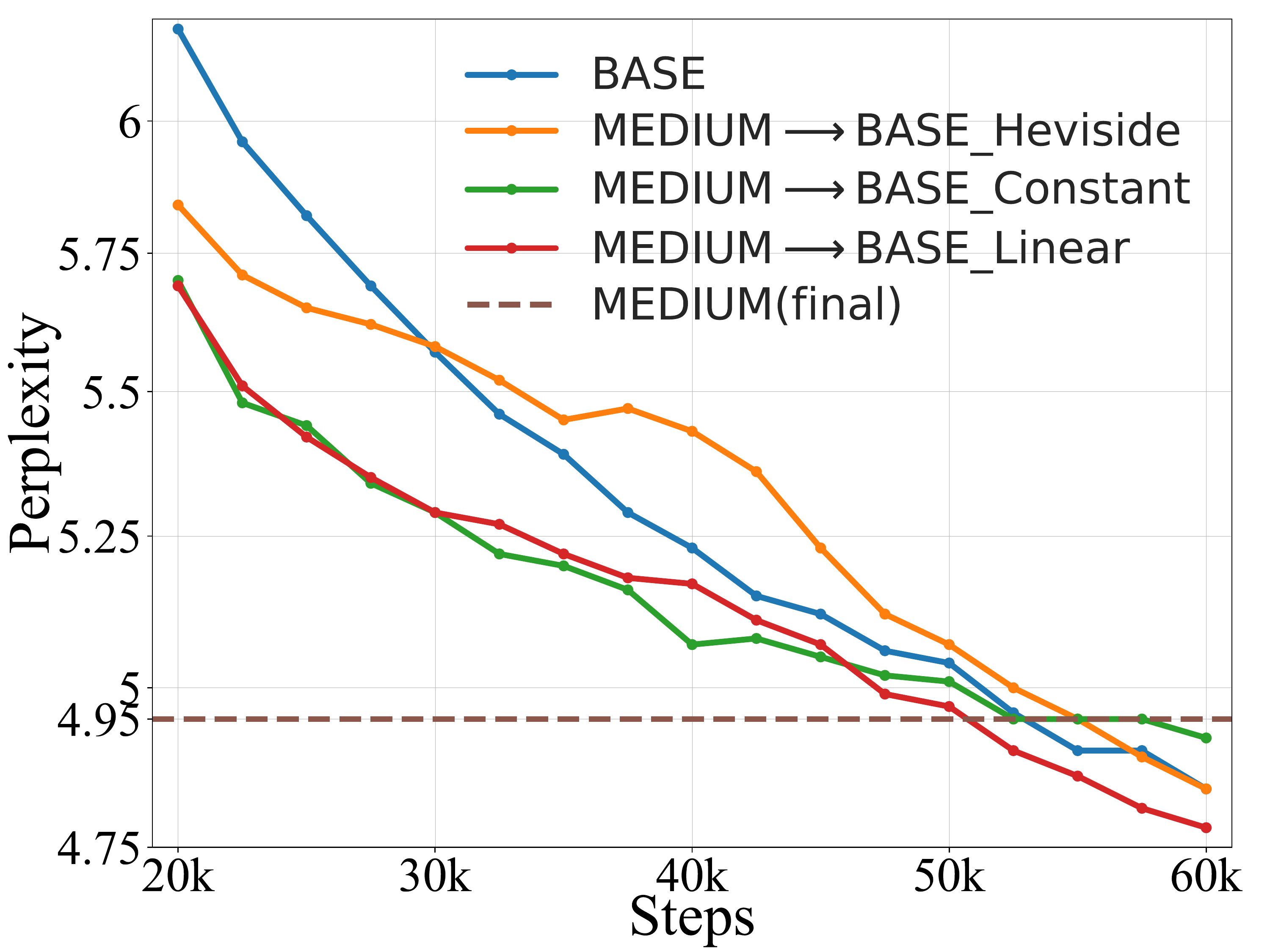}}
\centerline{\footnotesize{(b)}}
\end{minipage}
\hspace{0pt}
\begin{minipage}{0.3\textwidth}
\centerline{\includegraphics[width=\textwidth]{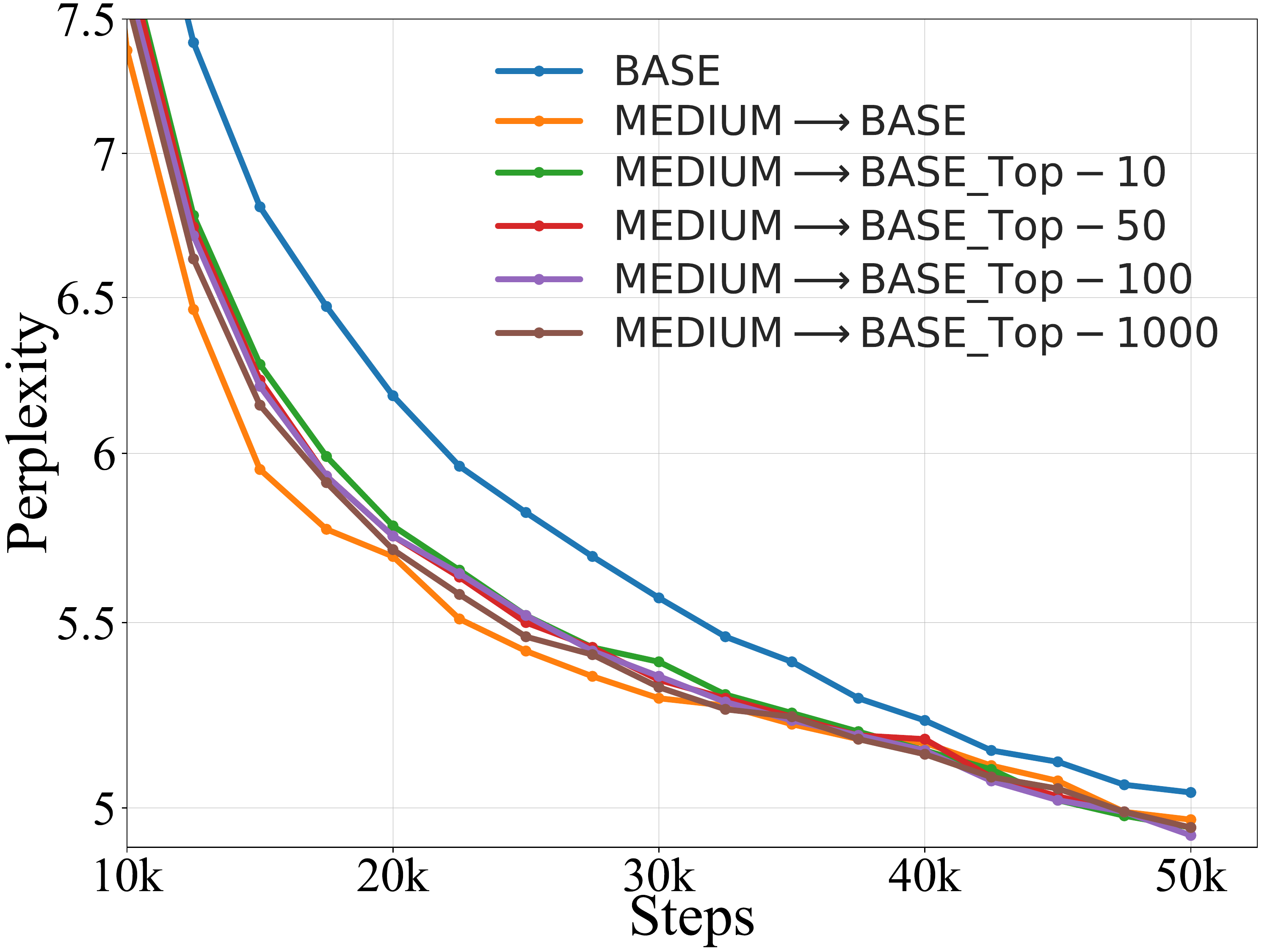}}
\centerline{\footnotesize{(c)}}
\end{minipage}
}
\caption{(a) The validation PPL curve for pre-training $\mathcal{M}_L$ under KI framework ($\texttt{BASE} \rightarrow \texttt{LARGE}$) and the self-learning baseline ($\texttt{LARGE}$). The teacher's ($\texttt{BASE}$) performance is $4.18$. (b) Pre-training $\texttt{BASE}$ under KI with three strategies for the inheritance rate $\alpha_t$: \texttt{Linear}, \texttt{Heviside} and \texttt{Constant}. The teacher's ($\texttt{MEDIUM}$) performance is $4.95$. (c) Pre-training $\texttt{BASE}$ under KI with top-$K$ logits, we vary $K$ in $\{10, 50, 100, 1000\}$, respectively.}
\label{fig:preliminary_time}
\end{figure*}

\subsection{RQ1: How Could Knowledge Inheritance Benefit Large PLMs' Training?}
\label{sec:prelim}
\paragraph{Setting.}
 Our KI framework is agnostic to the specific self-supervised pre-training task and the PLM architecture. Without loss of generality, we mainly focus on the representative MLM task and use the model architecture of RoBERTa~\cite{liu2019roberta}. Specifically, we first choose $\text{RoBERTa}_{\texttt{BASE}}$ (denoted as \texttt{BASE}) as the teacher ($\mathcal{M}_S$) and $\text{RoBERTa}_{\texttt{LARGE}}$ (denoted as \texttt{LARGE}) as the student ($\mathcal{M}_L$). We also experiment on auto-regressive language modeling using GPT~\cite{radford2018improving} to show KI is model-agnostic.
 
 For pre-training data, we use the concatenation of Wikipedia and BookCorpus~\cite{zhu2015aligning} same as BERT~\cite{devlin2018bert}, with roughly $3,400$M tokens in total. All models ($\mathcal{M}_S$ and $\mathcal{M}_L$) are trained for $125$k steps, with a batch size of $2,048$ and a sequence length of $512$. Note the whole training computations are comparable with those of BERT. 
 We pre-train $\mathcal{M}_L$ by inheriting $\mathcal{M}_S$'s knowledge under KI (denoted as ``$\texttt{BASE} \rightarrow \texttt{LARGE}$''). We compare it with ``$\texttt{LARGE}$'' that only conducts self-learning from beginning to end.

For performance evaluation, we report the validation perplexity (PPL) during pre-training and the downstream performance on development sets of eight GLUE~\cite{wang2018glue} tasks. Note compared with the self-learning baseline, in KI, the logits output by $\mathcal{M}_L$ are additionally used to calculate $\mathcal{L}_{\text{KI}}$, we empirically find that the additional computations caused by it are almost negligible compared with the cumbersome computations in Transformer blocks. Therefore, it requires almost the same computational cost between KI and the baseline for each step. Hence, we report the performance w.r.t training step~\cite{li2019budgeted}, while the performance w.r.t. FLOPs~\cite{schwartz2019green} and wall-clock time~\cite{li2020train} can be roughly obtained by stretching the figure horizontally.

\paragraph{Overall Results.}

As shown in Figure \ref{fig:preliminary_time} (a), we conclude that: (1) \textbf{training $\mathcal{M}_L$ under KI converges faster than the self-learning baseline}, indicating that inheriting the knowledge from an existing teacher is far more efficient than solely learning such knowledge. That is, to achieve the same level of validation PPL, KI requires fewer computational costs. Specifically, under the guidance of $\mathcal{M}_S$, whose validation PPL is $4.18$, $\texttt{BASE} \rightarrow \texttt{LARGE}$ achieves a validation PPL of $3.41$ at the end of pre-training, compared with baseline ($\texttt{LARGE}$) $3.58$. After $\texttt{BASE} \rightarrow \texttt{LARGE}$ stops learning from the teacher at the $40$k-th step, it improves the validation PPL from $4.60$ ($\texttt{LARGE}$) to $4.28$, which is almost the performance when the baseline $\texttt{LARGE}$ conducts self-learning for $55$k steps, thus saving roughly $27.3\%$ computational costs\footnote{If we load $\texttt{BASE}$ and compute its logits during pre-training, $18.7\%$ FLOPs can be saved roughly, since the forward passes of the small teacher also take up a small part.}. The results in Table \ref{tab:downstream} show that (2) \textbf{$\mathcal{M}_L$ trained under KI achieves better performance than the baseline on downstream tasks at each step}. We also found empirically that, under the same setting (e.g., data, hyper-parameters and model architectures), lower validation PPL generally indicates better downstream performance. Since the performance gain in downstream tasks is consistent with that reflected in PPL, we only show the latter for the remaining experiments. 
Concerning the energy cost, for the remaining experiments, unless otherwise specified, we choose $\texttt{MEDIUM}$ ($9$ layers, $576$ hidden size) as $\mathcal{M}_S$ and $\texttt{BASE}$ as $\mathcal{M}_L$.

\begin{table*}[!tbp]
  \centering
  \small
    \begin{tabular}{c@{~~~~~~}c@{~~~~~~}c@{~~~~~~}c@{~~~~~~}c@{~~~~~~}c@{~~~~~~}c@{~~~~~~}c@{~~~~~~}c@{~~~~~~}c@{~~~~~~}c}
    \toprule
    \textbf{Step}  & \textbf{Model} & \textbf{CoLA}  & \textbf{MNLI}  & \textbf{QNLI}  & \textbf{RTE}   & \textbf{SST-2} & \textbf{STS-B} & \textbf{MRPC}  & \textbf{QQP}   & \textbf{Avg} \\
    \midrule
    \multirow{2}[2]{*}{$5$k} & $\texttt{LARGE}$ & $0.0$   & $73.5$  & $81.7$  & $53.0$  & $81.7$  & $45.8$  & $71.4$  & $87.5$  & $61.8$  \\
          & $\texttt{BASE} \rightarrow \texttt{LARGE}$
          & $\mathbf{17.4}$ & $\mathbf{75.8}$ & $\mathbf{83.4}$ & $\mathbf{54.7}$ & $\mathbf{85.7}$ & $\mathbf{72.0}$ & $\mathbf{72.6}$ & $\mathbf{88.6}$ & $\mathbf{68.8}$ \\
    \midrule
    \multirow{2}[2]{*}{$45$k} & $\texttt{LARGE}$ & $61.8$  & $84.9$  & $91.7$  & $63.4$  & $92.9$  & $88.6$  & $87.7$  & $\mathbf{91.5}$ & $82.8$  \\
          & $\texttt{BASE} \rightarrow \texttt{LARGE}$ & $\mathbf{64.3}$ & $\mathbf{85.9}$ & $\mathbf{92.2}$ & $\mathbf{75.3}$ & $\mathbf{93.2}$ & $\mathbf{89.3}$ & $\mathbf{89.4}$ & $91.5$  & $\mathbf{85.2}$ \\
    \midrule
    \multirow{2}[2]{*}{$85$k} & $\texttt{LARGE}$ & $64.5$  & $86.8$  & $92.7$  & $69.7$  & $93.5$  & $89.9$  & $89.7$  & $91.7$  & $84.8$  \\
          & $\texttt{BASE} \rightarrow \texttt{LARGE}$ & $\mathbf{65.7}$ & $\mathbf{87.2}$ & $\mathbf{93.0}$ & $\mathbf{77.0}$ & $\mathbf{94.3}$ & $\mathbf{90.0}$ & $\mathbf{90.4}$ & $\mathbf{91.8}$ & $\mathbf{86.2}$ \\
    \midrule
    \multirow{2}[2]{*}{$125$k} & $\texttt{LARGE}$ & $64.3$  & $87.1$  & $\mathbf{93.2}$ & $73.4$  & $94.1$  & $90.3$  & $\mathbf{90.1}$ & $91.8$  & $85.5$  \\
          & $\texttt{BASE} \rightarrow \texttt{LARGE}$ & $\mathbf{67.7}$ & $\mathbf{87.7}$ & $93.1$  & $\mathbf{74.9}$ & $\mathbf{94.8}$ & $\mathbf{90.6}$ & $88.2$  & $\mathbf{91.9}$ & $\mathbf{86.1}$ \\
    \bottomrule
    \end{tabular}%
  \caption{Downstream performances on GLUE tasks (dev). KI requires fewer pre-training steps to get a high score after fine-tuning. Detailed results at different training steps are illustrated in \cref{sec:downstream_detail}.}
    \label{tab:downstream}
\end{table*}%




\paragraph{Effects of Inheritance Rate.}
We set $\alpha_t$ in Eq.~(\ref{KI_eq}) to be linearly decayed (denoted as \texttt{Linear}) to gradually encourage $\mathcal{M}_L$ exploring knowledge on its own. We analyze whether this design is necessary by comparing it with two other strategies: the first is to only learn from the teacher at first and change to pure self-learning (denoted as \texttt{Heviside}) at the $35$k-th step; the second is to use a constant ratio ($1:1$) between $\mathcal{L}_{\text{SELF}}$ and $\mathcal{L}_{\text{KI}}$ throughout the whole training process (denoted as \texttt{Constant}). We can conclude from Figure \ref{fig:preliminary_time} (b) that: (1) \textbf{annealing at first is necessary}. The validation PPL curve of \texttt{Linear} converges the fastest, while \texttt{Heviside} tends to increase after $\mathcal{M}_L$ stops learning from the teacher, indicating that, due to the difference between learning from the teacher and self-learning, annealing at first is necessary so that the performance won't decay at the transition point (the $35$k-th step). (2) \textbf{Supervision from the teacher is redundant after $\mathcal{M}_L$ surpasses $\mathcal{M}_S$}. Although \texttt{Constant} performs well in the beginning, its PPL gradually becomes even worse than the other two strategies. This indicates that after $\mathcal{M}_L$ has already surpassed $\mathcal{M}_S$, it will be encumbered by keeping following guidance from $\mathcal{M}_S$.

\paragraph{Saving Storage Space with Top-K Logits.} Loading the teacher $\mathcal{M}_S$ repeatedly for KI is cumbersome, and an alternative way is to pre-compute and save the predictions of $\mathcal{M}_S$ offline once and for all. We show that using the information of top-$K$ logits~\cite{tan2019multilingual} can reduce the memory footprint without much performance decrease. Specifically, we save only top-$K$ probabilities of $\mathcal{P}_S(x^j; \tau)$ followed by re-normalization, instead of the full distribution over all tokens. For RoBERTa, the dimension of $\mathcal{P}_S(x^j; \tau)$ is decided by its vocabulary size, which is around $50,000$. We thus vary $K$ in $\{10, 50, 100, 1000\}$ to see its effects in Figure \ref{fig:preliminary_time} (c), from which we observe that: \textbf{top-K logits contain the vast majority of information}. Choosing a relatively small $K$ (e.g., $10$) is already good enough for inheriting knowledge from the teacher without much performance decrease. Previous work also indicates the relation between KD and label smoothing~\cite{shen2021label}, however, we show in \cref{sec:exp_label_smoothing} that the improvements of KI are not because of benefiting from optimizing smoothed targets, which impose regularization.

\begin{figure*}[t]
\vfill
\centerline{
\begin{minipage}{0.3\textwidth}
    \centerline{\includegraphics[width=\textwidth]{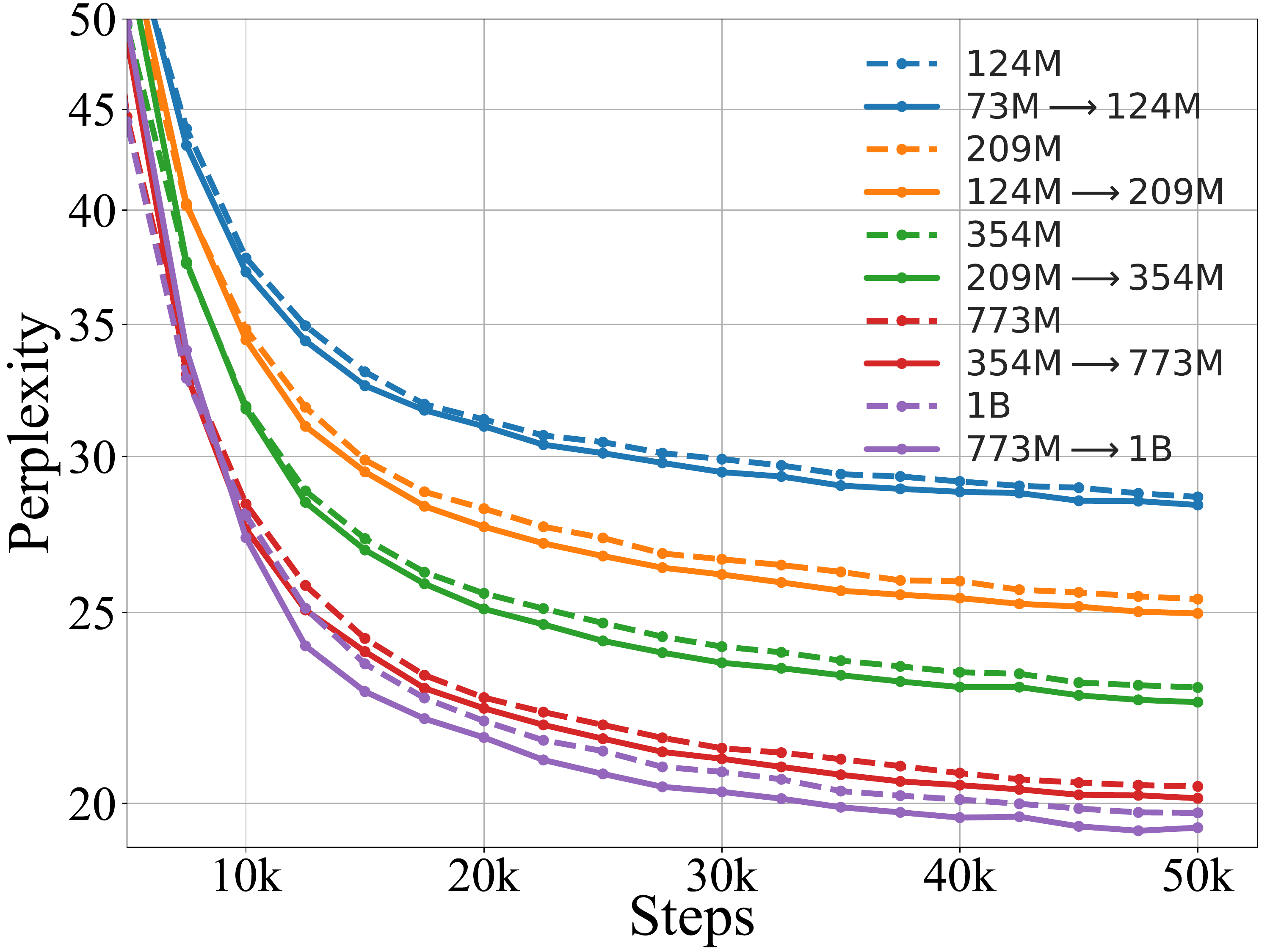}}
\centerline{\footnotesize{(a)}}
\end{minipage}
\hspace{0pt}
\begin{minipage}{0.3\textwidth}
    \centerline{\includegraphics[width=\textwidth]{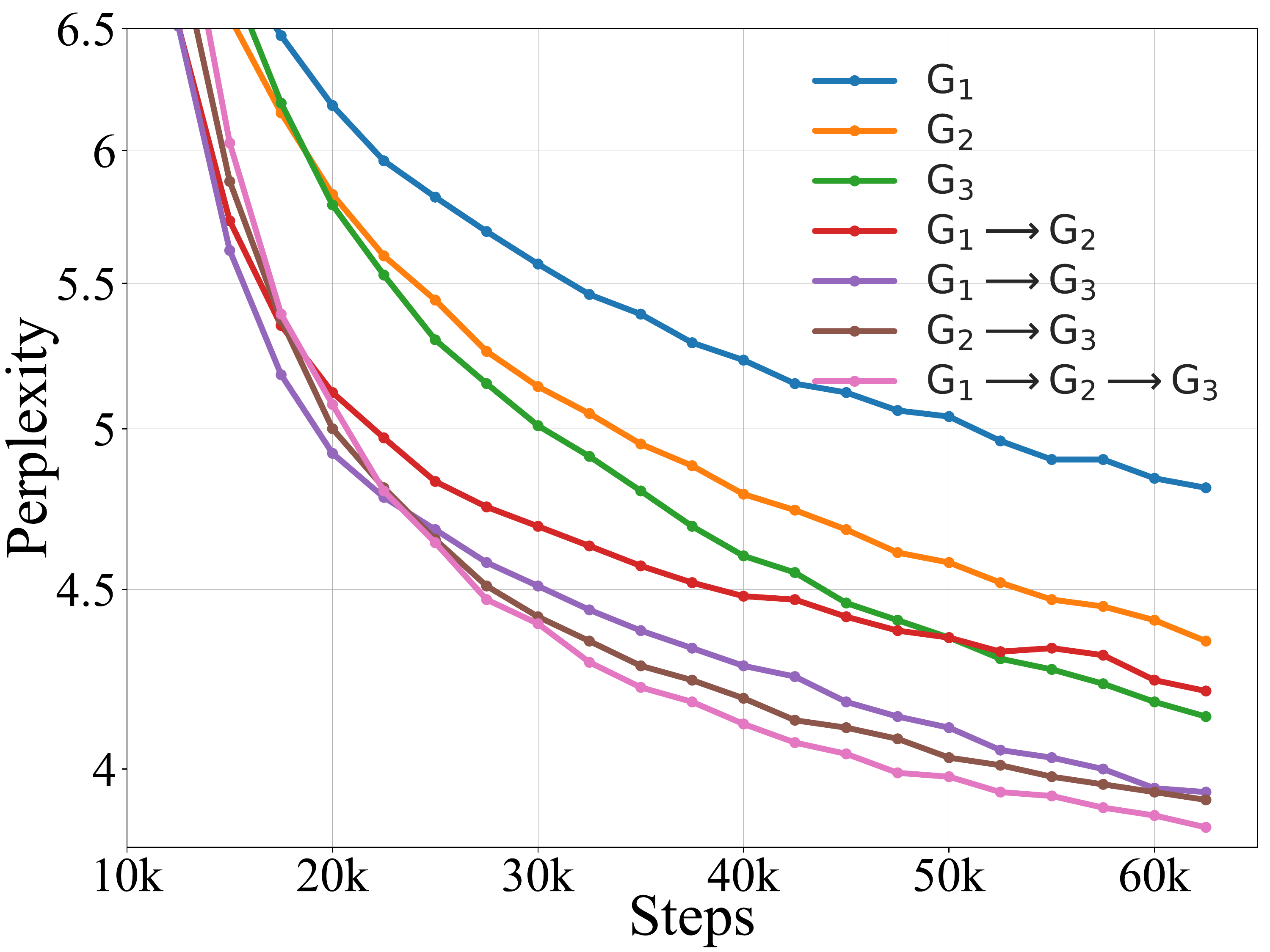}}
\centerline{\footnotesize{(b)}}
\end{minipage}
\hspace{0pt}
\begin{minipage}{0.3\textwidth}
\centerline{\includegraphics[width=\textwidth]{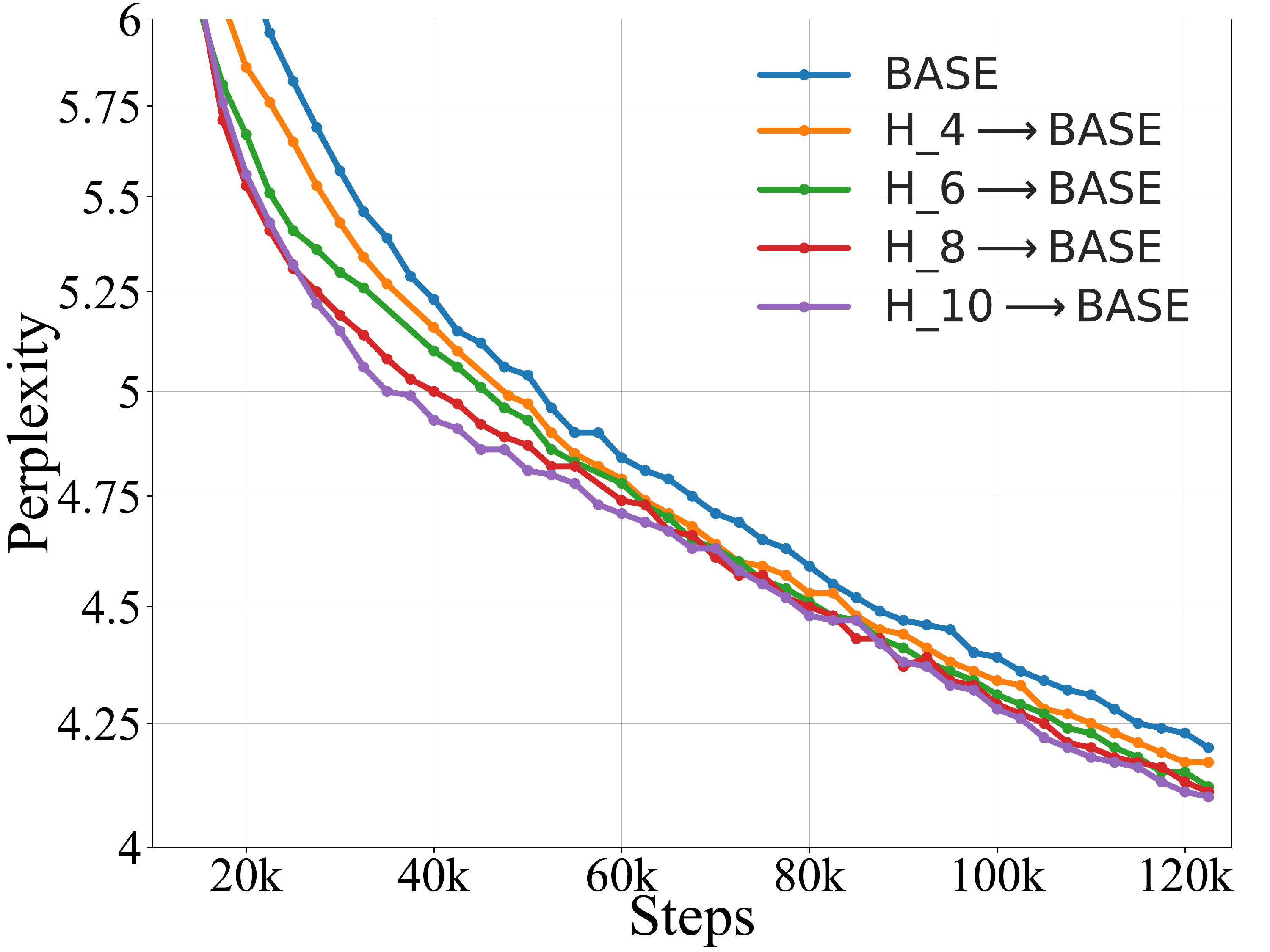}}
\centerline{\footnotesize{(c)}}
\end{minipage}
}
\caption{(a) Experiments on GPT. (b) KI over generations. (c) Effects of $\mathcal{M}_S$'s architecture (depth).}
\label{fig:gpt_generation_depth}
\end{figure*}

\paragraph{Experiments on GPT.} To demonstrate that KI is model-agnostic, we conduct experiments on auto-regressive language modeling and choose GPT~\cite{radford2018improving} architecture with growing sizes of $\{73\text{M}, 124\text{M}, 209\text{M}, 354\text{M}, 773\text{M},$ $1\text{B}\}$ parameters in total, respectively. The detailed architectures are specified in Table~\ref{tab:arch}. 
All the teacher models are pre-trained for $62.5$k steps with a batch size of $2,048$. As reflected in Figure \ref{fig:gpt_generation_depth} (a), training larger GPTs under our KI framework converges faster than the self-learning baseline, which demonstrates \textbf{KI is agnostic to the specific pre-training objective and PLM architecture}. 

\subsection{RQ2: Could Knowledge Inheritance be Performed over Generations?}
\label{generation}
Human beings can inherit the knowledge from their antecedents, refine it and pass it down to their offsprings, so that knowledge can gradually accumulate over generations. Inspired by this, we investigate whether PLMs also have this kind of pattern. Specifically, we experiment with the knowledge inheritance among three generations of RoBERTa with roughly $1.7$x growth in model size: $G_1$ ($\texttt{BASE}$, $125$M), $G_2$ ($\texttt{BASE\_PLUS}$, $211$M) and $G_3$ ($\texttt{LARGE}$, $355$M), whose architectures are listed in Table~\ref{tab:arch}. All models are trained from scratch for $125$k steps with a batch size of $2,048$ on the same corpus. We compare the differences among (1) self-learning for each generation (denoted as $G_1$, $G_2$ and $G_3$), (2) KI over two generations (denoted as $G_1 \rightarrow G_2$, $G_1 \rightarrow G_3$ and $G_2 \rightarrow G_3$), and (3) KI over three generations (denoted as $G_1 \rightarrow G_2 \rightarrow G_3$), where $G_2$ first inherits the knowledge from $G_1$, refines it by additional self-exploring and passes its knowledge down to $G_3$. The results are drawn in Figure \ref{fig:gpt_generation_depth} (b). Comparing the performance of $G_2$ and $G_1 \rightarrow G_2$, $G_3$ and $G_1 \rightarrow G_3$, or $G_3$ and $G_2 \rightarrow G_3$, we can again demonstrate the superiority of KI over self-training as concluded before. Comparing the performance of $G_1 \rightarrow G_3$ and $G_1 \rightarrow G_2 \rightarrow G_3$, or $G_2 \rightarrow G_3$ and $G_1 \rightarrow G_2 \rightarrow G_3$, it is observed that the performance of $G_3$ benefits from the involvements of both $G_1$ and $G_2$, which means knowledge could be accumulated through more generations' involvements.

\subsection{RQ3: How Could $\mathcal{M}_S$'s Pre-training Setting Affect Knowledge Inheritance?}
\label{sec:effects}

Existing PLMs are typically trained under quite different settings, and it is unclear how these different settings will affect the performance of KI. Formally, we have a series of well-trained smaller PLMs $\overline{\mathcal{M}_S}= \{\mathcal{M}_S^1,..., \mathcal{M}_S^{N_S}\}$, each having been optimized on $\overline{\mathcal{D}_S} = \{\mathcal{D}_S^1,..., \mathcal{D}_S^{N_S}\}$, respectively. Considering that the PLMs in $\overline{\mathcal{M}_S}$, consisting of varied model architectures, are pre-trained on different corpora of various sizes and domains with arbitrary strategies, thus the knowledge they master is also manifold. In addition, $\mathcal{M}_L$'s pre-training data $\overline{\mathcal{D}_L}$ may also consist of massive, heterogeneous corpora from multiple sources, i.e., $\overline{\mathcal{D}_L} = \{\mathcal{D}_L^1,..., \mathcal{D}_L^{N_L}\}$. Due to the difference between $\overline{\mathcal{D}_L}$ and $\overline{\mathcal{D}_S}$, $\mathcal{M}_S$ may be required to transfer its knowledge on instances unseen during its pre-training. Ideally, we want $\mathcal{M}_S$ to teach the courses it is skilled in. Therefore, it is essential to choose the most appropriate teacher for each composition $\mathcal{D}_L^* \in \overline{\mathcal{D}_L}$. To this end, we conduct thorough experiments to analyze the effects of several representative factors: model architecture, pre-training data, $\mathcal{M}_S$'s pre-training step (\cref{sec:exp_step}) and batch size (\cref{sec:exp_batch_size}).

\paragraph{Effects of Model Architecture.}
Large PLMs generally converge faster and achieve lower PPL, thus serving as more competent teachers. We experiment with two widely chosen architecture variations, i.e., depth (number of layers) and width (hidden size), to explore the effects of $\mathcal{M}_S$'s model architectures. We choose $\texttt{BASE}$ ($12$ layer, $768$ hidden size) as $\mathcal{M}_L$'s architecture, and choose the architecture of $\mathcal{M}_S$ to differ from $\mathcal{M}_L$ in either depth or width. Specifically, for $\mathcal{M}_S$, we vary the depth in $\{4, 6, 8, 10\}$, and the width in $\{384, 480, 576, 672\}$, respectively, and pre-train $\mathcal{M}_S$ under the same setting as $\mathcal{M}_L$. The PPL curve for each teacher model is shown in \cref{sec:exp_teacher_ppl}, from which we observe that deeper / wider teachers with more parameters converge faster and achieve lower PPL. After that, we pre-train $\mathcal{M}_L$ under KI leveraging these teacher models. As shown in Figure \ref{fig:gpt_generation_depth} (c) and \cref{sec:arch_hidden_size}, \textbf{choosing a deeper / wider teacher accelerates $\mathcal{M}_L$'s convergence}, demonstrating the benefits of learning from a more knowledgeable teacher. Since the performance of PLMs is weakly related to the model shape but highly related to the model size~\cite{li2020train}, it is always a better strategy to choose the larger teacher if other settings are kept the same. In experiments, we also find empirically that, the optimal duration of learning from the teacher is longer for larger teachers, which means it takes more time to learn from a more knowledgeable teacher.

\begin{figure*}[t]
\vfill
\centerline{
\begin{minipage}{0.3\textwidth}
\centerline{\includegraphics[width=\textwidth]{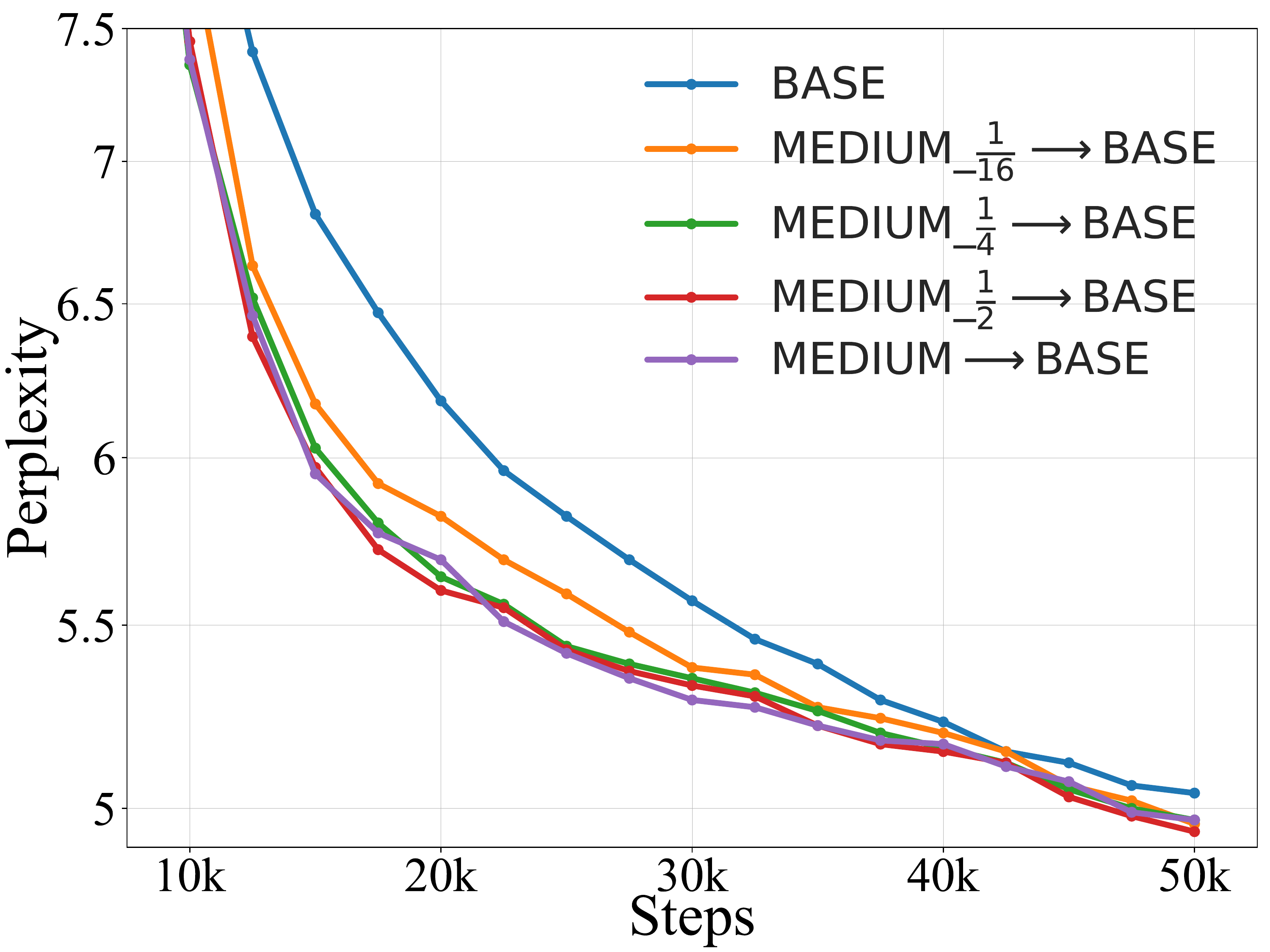}}
\centerline{\footnotesize{(a)}}
\end{minipage}
\hspace{0pt}
\begin{minipage}{0.3\textwidth}
\centerline{\includegraphics[width=\textwidth]{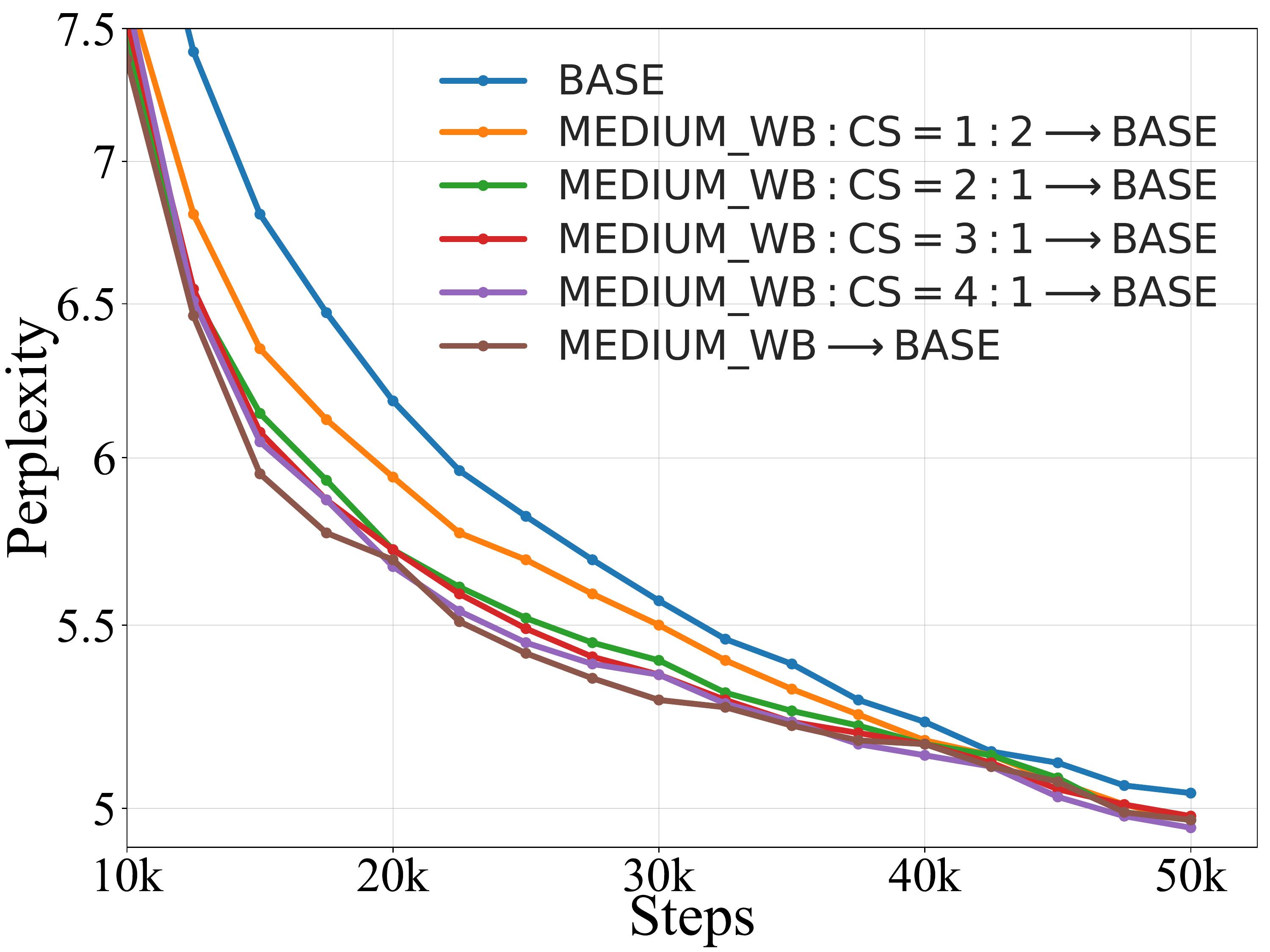}}
\centerline{\footnotesize{(b)}}
\end{minipage}
\hspace{0pt}
\begin{minipage}{0.3\textwidth}
\centerline{\includegraphics[width=\textwidth]{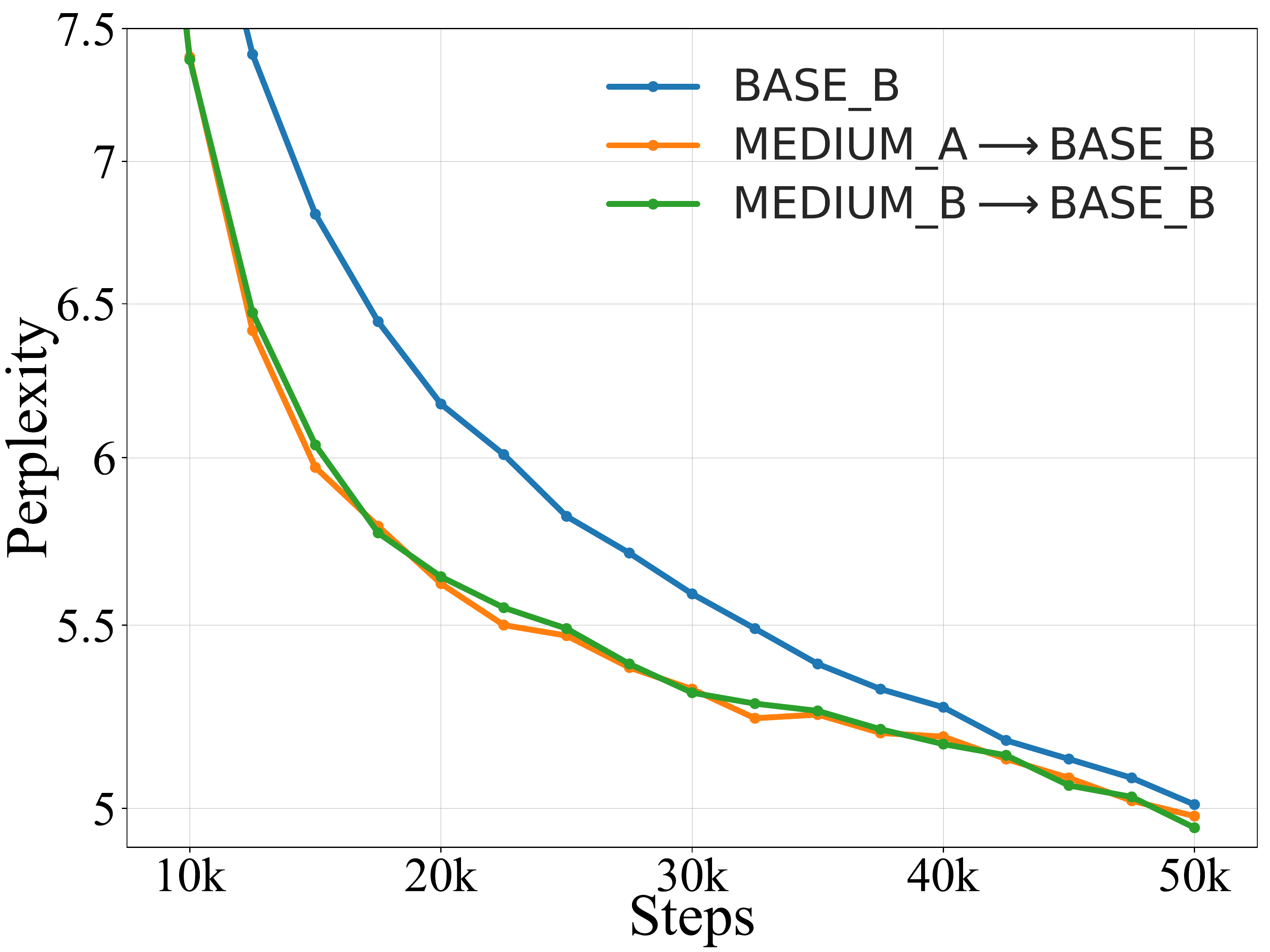}}
\centerline{\footnotesize{(c)}}
\end{minipage}
}
\caption{Effects of $\mathcal{M}_S$'s pre-training (a) data size, (b) data domain and (c) data privacy for KI.}
\label{fig:arch_size}
\end{figure*}

\paragraph{Effects of Pre-training Data.}
In previous experiments, we assume $\mathcal{M}_L$ is pre-trained on the same corpus as $\mathcal{M}_S$, i.e., $\mathcal{D}_L = \mathcal{D}_S$. However, in real-world scenarios, it may occur that the pre-training corpus used by both $\mathcal{M}_L$ and $\mathcal{M}_S$ is mismatched, due to three main factors: (1) \textbf{data size.} When training larger models, the pre-training corpus is often enlarged to improve downstream performance, i.e., $|\mathcal{D}_S| \ll |\mathcal{D}_L|$; (2) \textbf{data domain.} PLMs are trained on heterogeneous corpora from various sources (e.g., news articles, literary works, etc.), i.e., $\mathcal{P}_{\mathcal{D}_S} \neq \mathcal{P}_{\mathcal{D}_L}$. The different knowledge contained in each domain may affect PLMs' generalization in downstream tasks; (3) \textbf{data privacy.} Even if both size and domain of $\mathcal{D}_S$ and $\mathcal{D}_L$ are ensured to be the same, it may be hard to retrieve the pre-training corpus used by $\mathcal{M}_S$ due to privacy concerns, with an extreme case: $\mathcal{D}_L \cap \mathcal{D}_S = \emptyset$. The gap between $\mathcal{D}_S$ and $\mathcal{D}_L$ may hinder $\mathcal{M}_S$'s successful knowledge transfer. We thus design experiments to analyze the effects of these factors, with three observations concluded:

\textbf{$\bullet$ Obs. 1: PLMs can image the big from the small for in-domain data}. To evaluate the effects of data size, we first pre-train teacher models on different partitions of the original training corpus under the same setting by randomly sampling $\{\frac{1}{16}, \frac{1}{8}, \frac{1}{4}, \frac{1}{2}, \frac{1}{1}\}$ of it, resulting in teacher models with final validation PPL of $\{5.43,  5.15, 5.04, 4.98, 4.92\}$, respectively. The final validation PPL increases as we shrink the size of $\mathcal{M}_S$'s pre-training corpus, which implies that training with less data weakens the teacher's ability. Next, we compare the differences when their knowledge is inherited by $\mathcal{M}_L$. As reflected in Figure \ref{fig:arch_size} (a), however, the performance of KI is not substantially undermined until only $\frac{1}{16}$ of the original data is leveraged by the teacher. This indicates that PLMs can well image the overall data distribution even if it only sees a small part. Hence, when training larger PLMs, unless the data size is extensively enlarged, its impact can be ignored.


\textbf{$\bullet$ Obs. 2: Inheriting on similar domain improves performance}. To evaluate the effects of data domain, we experiment on the cases where $\mathcal{D}_S$ and $\mathcal{D}_L$ have domain mismatch. Specifically, keeping data size the same, we mix Wikipedia and BookCorpus (WB) used before with computer science (CS) papers from S2ORC~\cite{lo2019s2orc}, whose domain is distinct from WB, using different proportions, i.e., $\text{WB}: \text{CS} = \{1:2, 2:1, 3:1, 4:1\}$, respectively. We pre-train $\mathcal{M}_S$ on the constructed corpora, then test the performance when $\mathcal{M}_L$ inherits these teachers' knowledge on the WB domain data. As shown in Figure~\ref{fig:arch_size} (b), with the domain of the constructed corpus $\mathcal{M}_S$ is trained on becoming gradually similar to WB, the benefits from KI become more obvious, which means it is essential that both $\mathcal{M}_S$ and $\mathcal{M}_L$ are trained on similar domain of data, so that $\mathcal{M}_S$ can successfully impart knowledge to $\mathcal{M}_L$ by teaching the ``right'' course.

\textbf{$\bullet$ Obs. 3: Data privacy is not that important if the same domain is ensured}.  To evaluate the effects of data privacy, we experiment in an extreme case where $\mathcal{D}_S$ and $\mathcal{D}_L$ have no overlap at all. To avoid the influences of size and domain, we randomly split the WB domain training corpus $\mathcal{D}$ into two halves ($\mathcal{D}_{A}$ and $\mathcal{D}_{B}$) and pre-train two teacher models (denoted as $\texttt{MEDIUM}_\texttt{A}$ and $\texttt{MEDIUM}_\texttt{B}$) on them. After pre-training, both of them achieve almost the same final PPL ($4.99$) on the same validation set. They are then inherited by the student model $\texttt{BASE}$ on $\mathcal{D}_{B}$ (denoted as $\texttt{MEDIUM}_\texttt{A} \rightarrow \texttt{BASE}_\texttt{B}$ and $\texttt{MEDIUM}_\texttt{B} \rightarrow \texttt{BASE}_\texttt{B}$), which is exactly the pre-training corpus of $\texttt{MEDIUM}_\texttt{B}$ and has no overlap with that of $\texttt{MEDIUM}_\texttt{A}$. We also choose $\mathcal{M}_L$ that conducts pure self-learning on $\mathcal{D}_B$ as the baseline (denoted as $\texttt{BASE}_\texttt{B}$). It is observed from Figure \ref{fig:arch_size} (c) that, there is little difference between the validation PPL curves of $\texttt{MEDIUM}_\texttt{A} \rightarrow \texttt{BASE}_\texttt{B}$ and $\texttt{MEDIUM}_\texttt{B} \rightarrow \texttt{BASE}_\texttt{B}$, indicating that whether the pre-training corpus of $\mathcal{M}_S$ and $\mathcal{M}_L$ has data overlap or not is not a serious issue as long as they share the same domain. This is meaningful when organizations aim to share the knowledge of their PLMs without exposing either the pre-training data or the model parameters due to privacy concerns. 

\subsection{RQ4: How Could Knowledge Inheritance Benefit Domain Adaptation?}
\label{sec:continual}

\begin{table*}[t]
  \centering
  \small
    \begin{tabular}{llcccccccccc}
    \toprule
    \multicolumn{2}{l}{$\textbf{N}_{\textbf{tokens}}$} & \multicolumn{2}{c}{$3,400$M} & \multicolumn{2}{c}{$200$M} & \multicolumn{2}{c}{$100$M} & \multicolumn{2}{c}{$40$M}  &
    \multicolumn{2}{c}{$20$M}  \\
    \midrule
    \multicolumn{2}{l}{\textbf{Metrics}} & F1   & PPL   & F1   & PPL   & F1   & PPL   & F1   & PPL & F1   & PPL \\
    \midrule
    \multirow{2}[2]{*}{CS} & \text{SL}    & $69.8$ & $3.12$  & $71.7$ & $3.17$  & $71.4$ & $3.24$  & $68.3$ & $3.51$ & $67.5$ & $4.07$ \\
          & \text{KI}    & $\mathbf{72.9}$ & $\mathbf{3.06}$ & $\mathbf{72.6}$ & $\mathbf{3.09}$ & $\mathbf{71.9}$ & $\mathbf{3.11}$ & $\mathbf{71.1}$ & $\mathbf{3.21}$ &
          $\mathbf{70.8}$ &  $\mathbf{3.37}$
          \\
    \midrule
    \multirow{2}[2]{*}{BIO} & \text{SL}    & $84.0$ & $2.67$  & $82.8$ & $2.72$  & $83.2$ & $2.83$  & $83.3$ & $3.16$ & $82.7$ & $3.81$ \\
          & \text{KI}    & $\mathbf{84.5}$ & $\mathbf{2.65}$ & $\mathbf{83.4}$ & $\mathbf{2.66}$ & $\mathbf{83.9}$ & $\mathbf{2.69}$ & $\mathbf{83.6}$ & $\mathbf{2.82}$ &
          $\mathbf{83.5}$ &  $\mathbf{3.01}$ \\
    \bottomrule
    \end{tabular}%
    \caption{The validation PPL ($\text{PPL}$) and downstream performance (F1) on the target domain (CS / BIO) after $\texttt{BASE}_{\texttt{WB}}$ is post-trained for $4$k steps with self-learning (SL) or knowledge inheritance (KI). We experiment with different sizes of domain corpus. All downstream experiments are repeated $10$ times with different seeds.}
    \label{tab:continual}
\end{table*}%

\begin{table*}[!t]
  \centering
  \small
    \begin{tabular}{l@{~~~}c@{~~~}c@{~~~}c@{~~~}c@{~~~}c@{~~~}c@{~~~}c@{~~~}c@{~~~}c@{~~~}c@{~~~}c@{~~~}c@{~~~}c@{~~~}c@{~~~}c@{~~~}c}
    \toprule
    $\textbf{N}_{\textbf{tokens}}$ & \multicolumn{4}{c}{$3,400$M}       & \multicolumn{4}{c}{$200$M}        & \multicolumn{4}{c}{$100$M}      & \multicolumn{4}{c}{$40$M} \\
    \midrule
    \textbf{Metrics} & $\text{F1}_{\text{C}}$    & $\text{PPL}_{\text{C}}$   & $\text{F1}_{\text{B}}$    & $\text{PPL}_{\text{B}}$   & $\text{F1}_{\text{C}}$    & $\text{PPL}_{\text{C}}$   & $\text{F1}_{\text{B}}$    & $\text{PPL}_{\text{B}}$   & $\text{F1}_{\text{C}}$    & $\text{PPL}_{\text{C}}$   & $\text{F1}_{\text{B}}$    & $\text{PPL}_{\text{B}}$   & $\text{F1}_{\text{C}}$    & $\text{PPL}_{\text{C}}$   & $\text{F1}_{\text{B}}$    & $\text{PPL}_{\text{B}}$ \\
    \midrule
    SL    & $71.7$ & $3.15$  & $83.7$ & $2.71$  & $70.5$ & $3.97$  & $82.7$ & $3.36$  & $67.7$ & $5.95$  & $81.7$ & $4.84$  &   $68.3$    & $11.7$ &   $81.1$    & $10.5$ \\
    KI    & $\mathbf{72.2}$ & $\mathbf{3.15}$  & $\mathbf{83.9}$ & $\mathbf{2.70}$   & $\mathbf{71.8}$ & $\mathbf{3.42}$  & $\mathbf{83.1}$ & $\mathbf{2.92}$  & $\mathbf{69.8}$ & $\mathbf{3.90}$   & $\mathbf{82.6}$ & $\mathbf{3.32}$  &  $\mathbf{69.1}$     & $\mathbf{5.70}$   &   $\mathbf{81.3}$    & $\mathbf{4.64}$ \\
    \bottomrule
    \end{tabular}%
  \caption{The results when $\texttt{BASE}_{\texttt{WB}}$ is post-trained on two new domains simultaneously with self-learning (SL) or knowledge inheritance (KI). We report both validation PPL ($\text{PPL}_{\text{B}}$ / $\text{PPL}_{\text{C}}$) and downstream performance ($\text{F1}_{\text{B}}$ / $\text{F1}_{\text{C}}$) for BIO / CS domain. We observe that SL exhibits severe overfitting when data is relatively scarce.}
  \label{tab:continual_combine}%
\end{table*}%

With streaming data of various domains continuously emerging, training domain-specific PLMs and storing the model parameters for each domain can be prohibitively expensive. To this end, researchers recently demonstrated the feasibility of adapting PLMs to the target domain through continual pre-training~\cite{gururangan2020don}. In this section, we further extend KI and demonstrate that domain adaptation for PLM can benefit from inheriting knowledge of existing domain experts.

Specifically, instead of training large PLMs from scratch, which is the setting used before, we focus on adapting $\texttt{BASE}_{\texttt{WB}}$, which has been well-trained on the WB domain for $125$k steps, to two target domains, i.e., computer science (CS) and biomedical (BIO) papers from S2ORC~\cite{lo2019s2orc}. The proximity (vocabulary overlap) of three domains is listed in \cref{sec:domain_proximity}. We assume there exist two domain experts, i.e., $\texttt{MEDIUM}_{\texttt{CS}}$ and $\texttt{MEDIUM}_{\texttt{BIO}}$. Each model has been trained on CS / BIO domain for $125$k steps. Note their training computation is far less than $\texttt{BASE}_{\texttt{WB}}$ due to fewer model parameters. Hence, either $\texttt{MEDIUM}_{\texttt{CS}}$ or $\texttt{MEDIUM}_{\texttt{BIO}}$ is no match for $\texttt{BASE}_{\texttt{WB}}$ in WB domain but has richer knowledge in CS / BIO domain. For evaluation, we compare both (1) the validation PPL on the target domain and (2) the performance (test F1) on downstream tasks, i.e. ACL-ARC~\cite{jurgens2018measuring} for CS domain and CHEMPROT~\cite{kringelum2016chemprot} for BIO domain. Before adaptation, $\texttt{BASE}_{\texttt{WB}}$ achieves a PPL of $5.41$ / $4.86$ and F1 of $68.5$ / $81.6$ on CS / BIO domain, while $\texttt{MEDIUM}_{\texttt{CS}}$ achieves $2.95$ (PPL) and $69.4$ (F1) on CS domain, $\texttt{MEDIUM}_{\texttt{BIO}}$ achieves $2.55$ (PPL) and $83.6$ (F1) on BIO domain. This demonstrates the superiority of two teachers over the student in their own domain despite their smaller model capacity.

We compare two strategies for domain adaptation: (1) only conducting self-learning on the target domain and (2) inheriting knowledge from well-trained domain teachers. Specifically, $\texttt{BASE}_{\texttt{WB}}$ is post-trained for additional $4$k steps on either CS or BIO domain to learn new knowledge. In addition, considering that in real-world scenarios, it can be hard to retrieve enough pre-training data for a specific domain, due to some privacy issues. Hence, we conduct experiments with different sizes of domain corpus. In Table \ref{tab:continual}, $\texttt{BASE}_{\texttt{WB}}$ is post-trained on either CS or BIO domain while in Table \ref{tab:continual_combine}, it is trained on synthetic domain data ($\text{BIO}:\text{CS}$ = $1:1$) to absorb knowledge from two domains simultaneously (we assume $\mathcal{M}_L$ is trained with the optimal teacher selection strategy, i.e., each teacher imparts the knowledge on its own domain data). It can be concluded from Table \ref{tab:continual} and Table \ref{tab:continual_combine} that:

(1) \textbf{KI is more training-efficient}. Compared with self-learning, inheriting knowledge from domain teachers achieves lower final PPL and improved performance in domain-specific downstream tasks, indicating that, for domain adaptation, KI is more training-efficient so that an already trained large PLM could absorb more knowledge from new domain with the same training budget. (2) \textbf{KI is more data-efficient}. The PPL gap between KI and SL is further enlarged when there is less domain-specific data available for adaptation, which means KI is more data-efficient especially under the low-resource setting, where domain data is scarce. In other words, only providing a small portion of domain-specific data is enough for satisfactory adaptation performance under KI, while self-learning exhibits overfitting to some extent. (3) \textbf{Large PLMs can simultaneously absorb knowledge from multiple domains and thus become omnipotent}. From Table \ref{tab:continual_combine}, we observe $\texttt{BASE}_{\texttt{WB}}$ achieves improved performance on both domains after being taught by two teachers simultaneously. KI shows superiority over self-learning. However, simultaneous learning overfits training data more easily and its performance on either domain is no match for learning only one domain at a time. 
\section{Conclusion and Future Work}
In this work, we propose a general knowledge inheritance (KI) framework that leverages previously trained PLMs for training larger ones. We conduct sufficient empirical studies to demonstrate its feasibility. In addition, we show that KI could well support knowledge transfer over a series of PLMs with growing sizes. We also comprehensively analyze various pre-training settings of the teacher model that may affect KI's performance, the results shed light on how to choose the most appropriate teacher PLM for KI. Finally, we extend KI and show that, during domain adaptation, an already trained large PLM could benefit from smaller domain teachers. In general, we provide a promising direction to share and exchange the knowledge learned by different models and continuously promote their performance.

In future, we aim to explore the following directions: (1) the \textbf{efficiency} of KI, i.e., given limited computational budget and pre-training corpus, how to more efficiently absorb knowledge from teacher models. Potential solutions include denoising teacher models' predictions and utilizing more information from the teacher. How to select the most representative data points for KI is also an interesting topic; (2) the \textbf{effectiveness} of KI under different settings, i.e., how can KI be applied if the teachers and the students are pre-trained on different vocabularies, languages, pre-training objectives and modalities.

Finally, we believe it is vital to use fair benchmarking that can accurately and reliably judge each KI algorithm. Thus, we suggest future work to: (1) conduct all experiments under the same computation environment and report the pre-training hyper-parameters and hardware deployments in detail, (2) evaluate the downstream tasks with multiple different random seeds and choose tasks that give relatively stable and consistent results, which could serve as better indicators for PLMs’ effectiveness. In addition, it is also essential that PLMs are tested on diverse downstream tasks which evaluate PLMs’ different abilities, (3) save the checkpoint more frequently during pre-training and evaluate the downstream performance, which can better indicate the trend of PLMs’ effectiveness, and (4) open-source all the codes and model parameters for future comparisons.

\section*{Acknowledgments}
This work is supported by the National Key R\&D Program of China (No. 2020AAA0106502), Institute Guo Qiang at Tsinghua University, NExT++ project from the National Research Foundation, Prime Minister’s Office, Singapore under its IRC@Singapore Funding Initiative, and Beijing Academy of Artificial Intelligence (BAAI). This work is also supported by the Pattern Recognition Center, WeChat AI, Tencent Inc. Yujia Qin and Yankai Lin designed the methods and the experiments. Yujia Qin, Jing Yi and Jiajie Zhang conducted the experiments. Yujia Qin, Yankai Lin and Xu Han wrote the paper. Zhiyuan Liu, Peng Li, Maosong Sun and Jie Zhou advised the project. All the authors participated in the discussion.

\bibliography{anthology,custom}
\bibliographystyle{acl_natbib}

\clearpage
\appendix
\section*{Appendices}

\section{Additional Experiments and Analysis}

\subsection{Effects of Model Size}
\label{sec:exp_model_size}
We experiment on four PLMs with roughly $1.7$x growth in model size: $\mathcal{M}_1$ ($\text{RoBERTa}_{\texttt{MEDIUM}}$, $73.5$M), $\mathcal{M}_2$ ($\text{RoBERTa}_{\texttt{BASE}}$, $125$M), $\mathcal{M}_3$ ($\text{RoBERTa}_{\texttt{BASE\_PLUS}}$, $211$M) and $\mathcal{M}_4$ ($\text{RoBERTa}_{\texttt{LARGE}}$, $355$M), whose architectures are listed in Table \ref{tab:arch}. We first pre-train a teacher PLM $\mathcal{M}_i$ ($\mathcal{M}_S$) for $125$k steps with a batch size of $2,048$ under the same setting then train a larger one $\mathcal{M}_{i+1}$ ($\mathcal{M}_L$) by inheriting $\mathcal{M}_i$'s knowledge under KI framework (denoted as $\mathcal{M}_{i} \rightarrow \mathcal{M}_{i+1}, i \in \{1,2,3\}$). We compare $\mathcal{M}_{i} \rightarrow \mathcal{M}_{i+1}$ with $\mathcal{M}_{i+1}$ that conducts self-learning from beginning to end. As shown in Figure \ref{fig:size_time_batch}, the superiority of KI is observed across all models. In addition, with the overall model size of $\mathcal{M}_S$ and $\mathcal{M}_L$ gradually increasing, the benefits of KI become more evident, reflected in the broader absolute gap between the PPL curve of $\mathcal{M}_{i} \rightarrow \mathcal{M}_{i+1}$ and $\mathcal{M}_{i+1}$ when $i$ gradually grows. This implies that with the advance of computing power in future, training larger PLMs will benefit more and more from our KI framework.

\subsection{Effects of $\mathcal{M}_S$'s Pre-training Steps}
\label{sec:exp_step}
\begin{figure*}[h]
    \centering
    \subfigure{\includegraphics[width=0.3\textwidth]{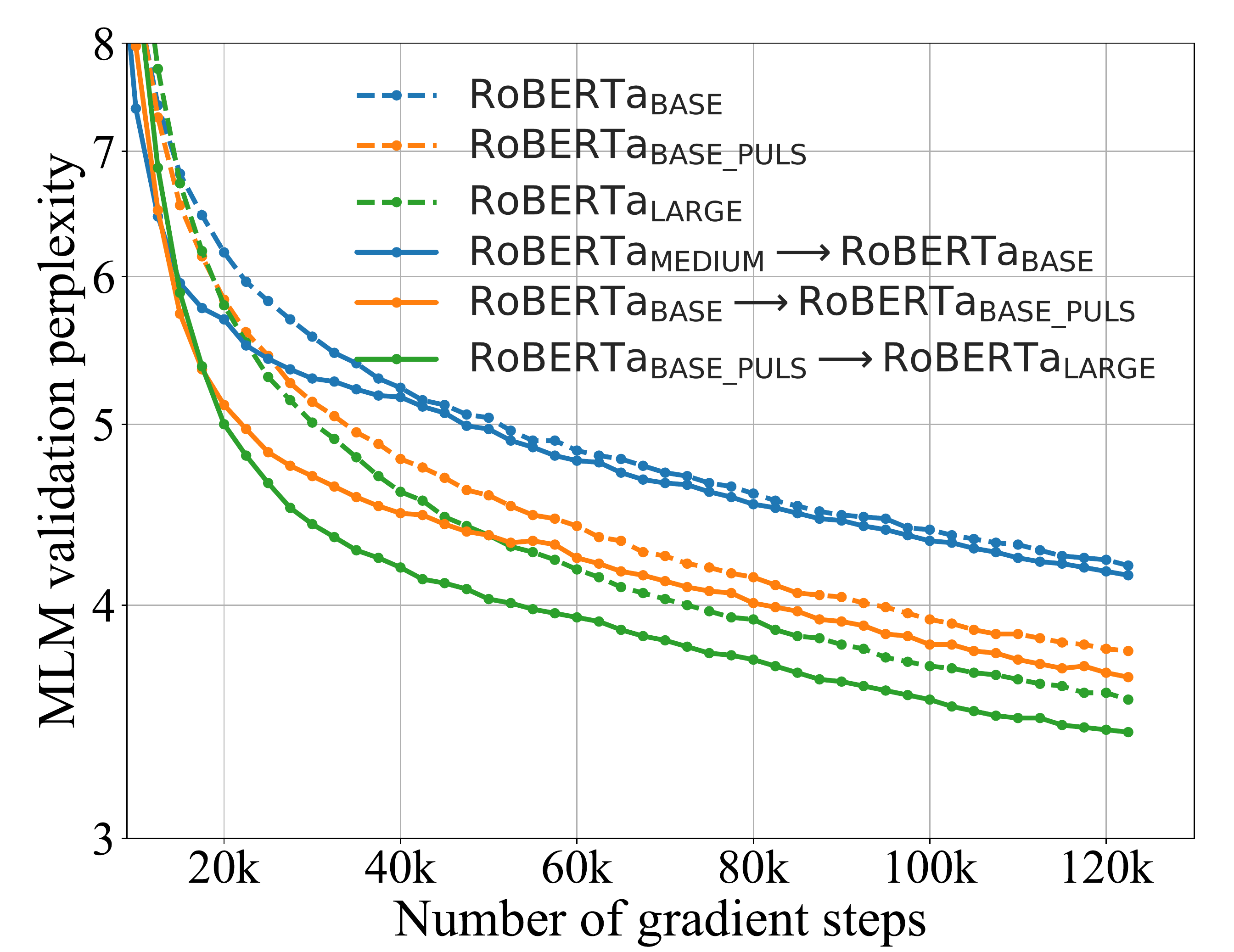}} 
    \subfigure{\includegraphics[width=0.3\textwidth]{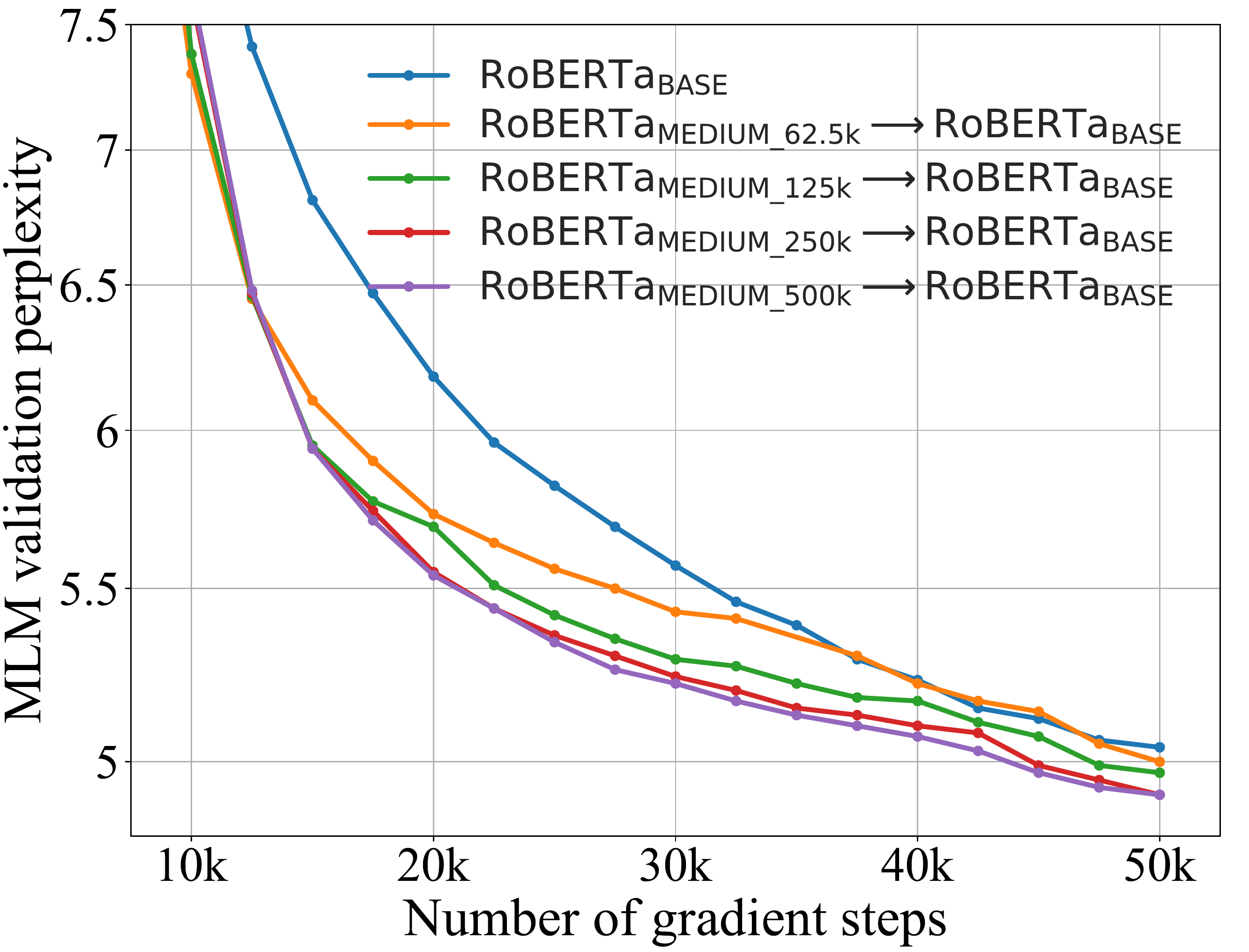}} 
    \subfigure{\includegraphics[width=0.3\textwidth]{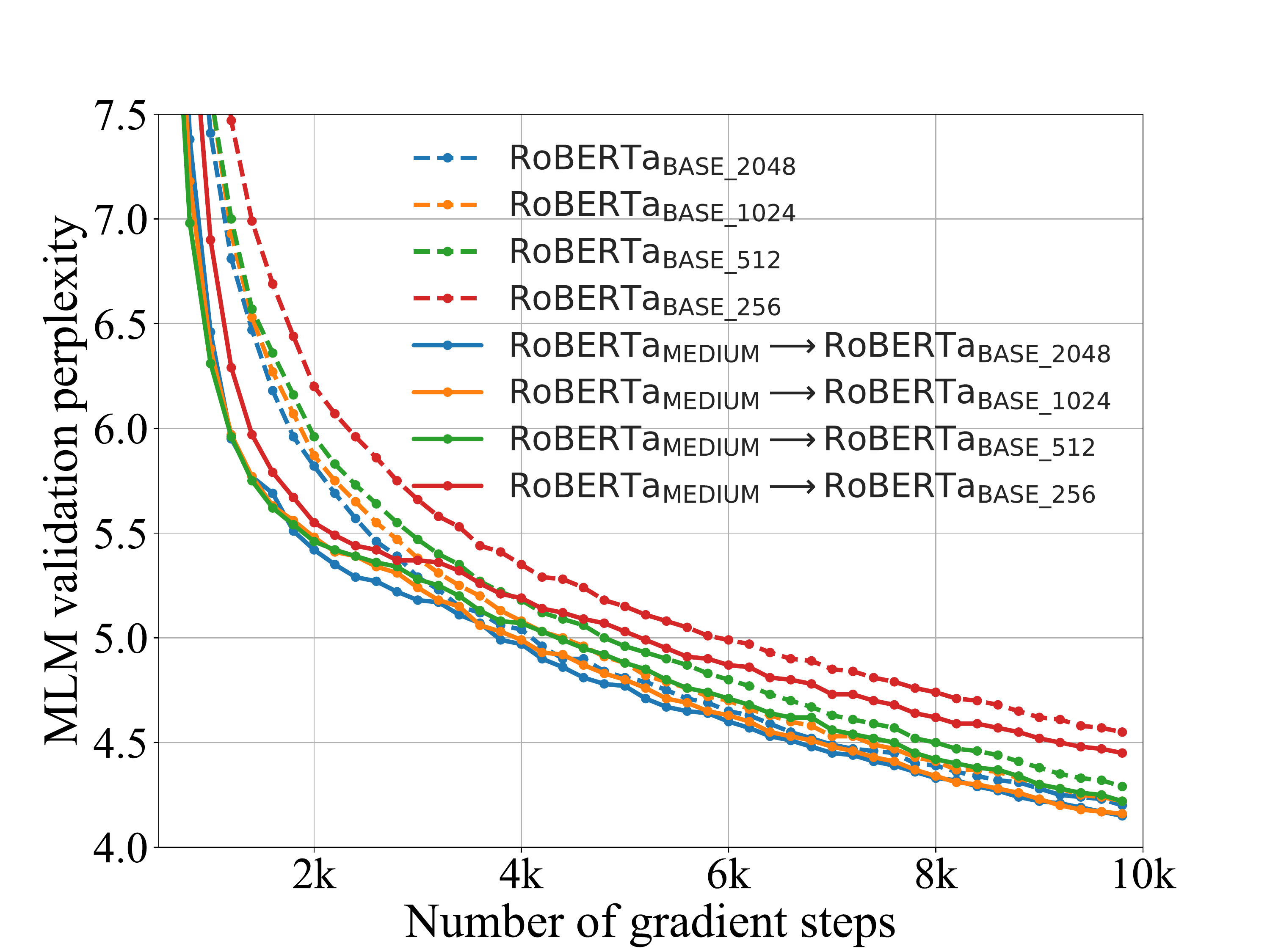}} 
    \caption{Left: effects of $\mathcal{M}_L$'s model size. Middle: effects of $\mathcal{M}_S$'s number of pre-training steps. Right: effects of $\mathcal{M}_L$'s batch size.}
    \label{fig:size_time_batch}
\end{figure*}

Longer pre-training has been demonstrated as an effective way for PLMs to achieve better performance~\cite{liu2019roberta} and thus become more knowledgeable. To evaluate the benefits of more pre-training steps for $\mathcal{M}_S$, we first vary $\text{RoBERTa}_{\texttt{MEDIUM}}$'s pre-training steps in $\{62.5$k, $125$k, $250$k, $500\text{k}\}$, and keep all other settings the same. After pre-training, these teacher models achieve the final validation PPL of $\{5.25, 4.92, 4.72, 4.51\}$, respectively. Then we compare the performances when $\text{RoBERTa}_{\texttt{BASE}}$ learn from these teacher models and visualize the results in Figure \ref{fig:size_time_batch}, from which we can conclude that, inheriting knowledge from teachers with longer pre-training time (steps) helps $\mathcal{M}_L$ converge faster. However, such a benefit is less and less obvious as $\mathcal{M}_S$'s pre-training steps increase, which means after enough training computations are invested, the teacher model enters a plateau of convergence in validation PPL, and digging deeper in knowledge becomes even harder. The bottleneck lies in other factors, e.g., the size and diversity of pre-training data, which hinder $\mathcal{M}_S$ from becoming more knowledgeable. We also found empirically that, after being pre-trained for $125$k steps on the corpus with a batch size of $2,048$, all the models used in this paper have well converged, and longer pre-training only results in limited performance gain in either PPL or downstream performance.



\begin{figure*}[h]
    \centering
    \subfigure{\includegraphics[width=0.3\textwidth]{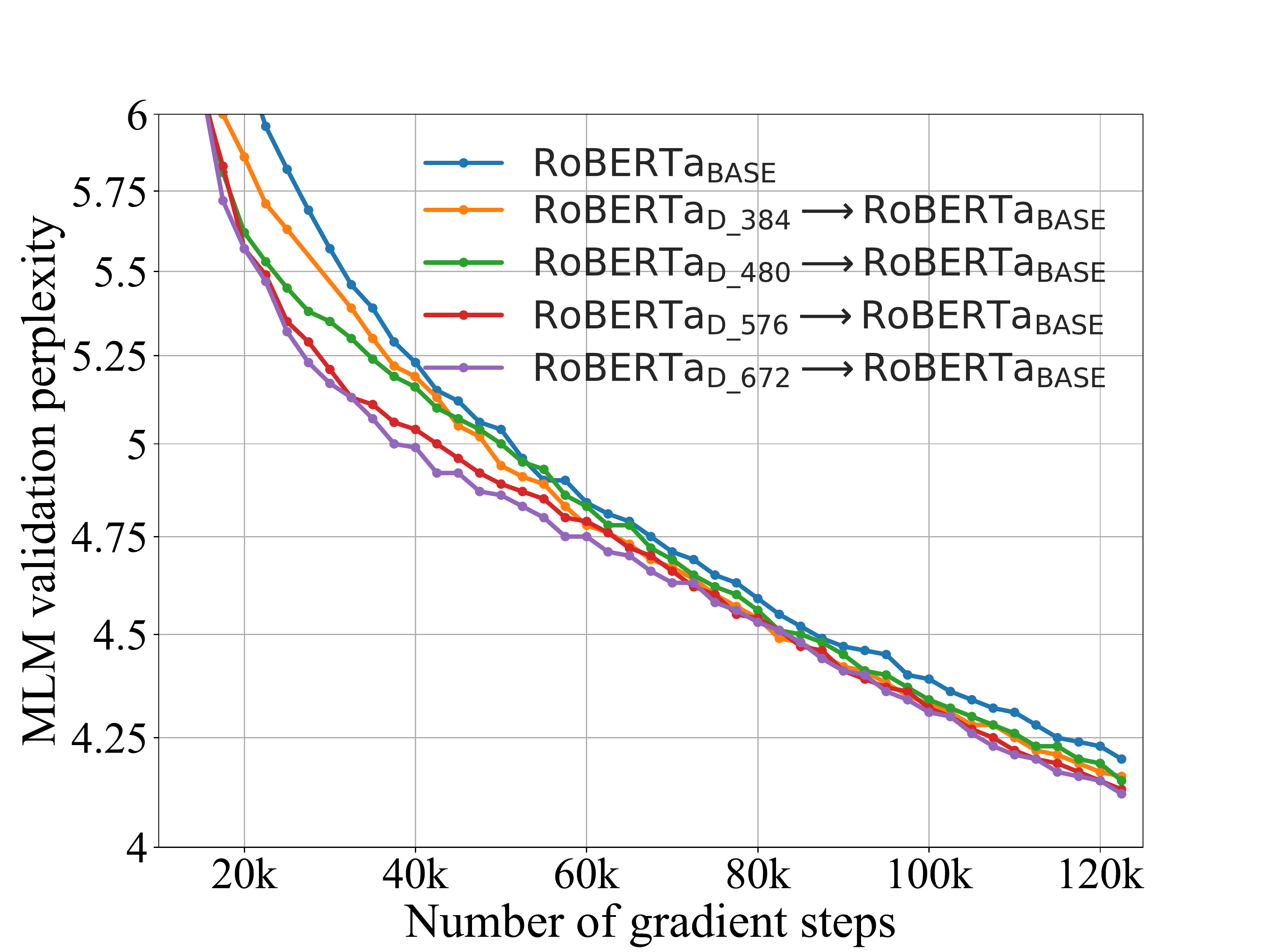}} 
    \subfigure{\includegraphics[width=0.3\textwidth]{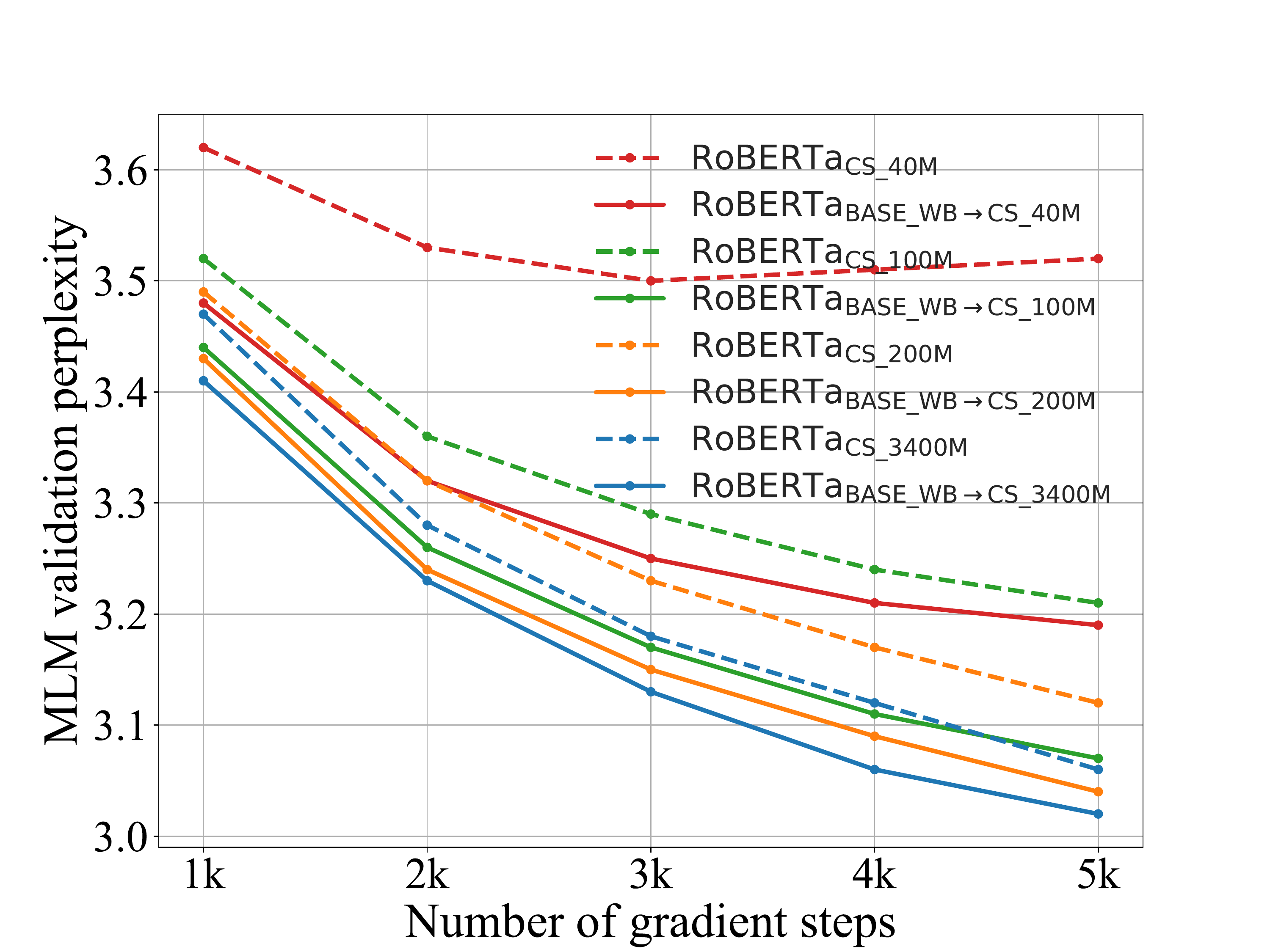}} 
    \subfigure{\includegraphics[width=0.3\textwidth]{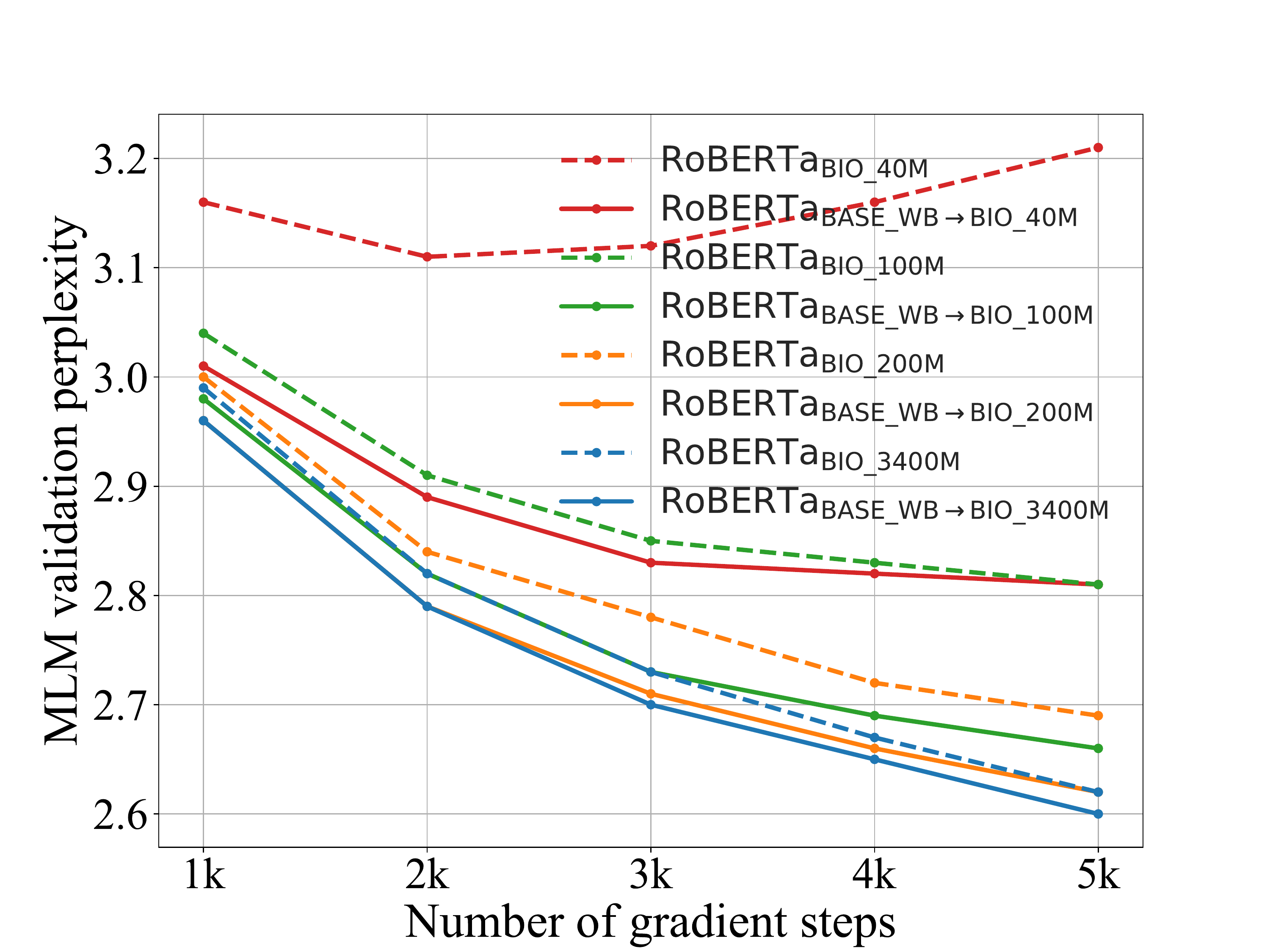}} 
    \caption{Left: the PPL curve when choosing the teacher PLM with different hidden sizes. Middle \& Right: adapting $\text{RoBERTa}_{\texttt{BASE\_WB}}$ to CS (middle) / BIO (right) domain with different number of training steps on different sizes of domain data. We compare two strategies: self-learning and KI. For example, $\text{RoBERTa}_{\texttt{CS\_3400M}}$ denotes post-training $\text{RoBERTa}_{\texttt{BASE\_WB}}$ with the self-learning strategy on the $3,400$M token CS domain corpus. $\text{RoBERTa}_{\texttt{BASE\_WB} \rightarrow \texttt{CS\_3400M}}$ denotes post-training $\text{RoBERTa}_{\texttt{BASE\_WB}}$ with the KI strategy on the $3,400$M token CS domain corpus.}
    \label{fig:gpt_continual}
\end{figure*}

\subsection{Effects of $\mathcal{M}_L$'s Batch Size}
\label{sec:exp_batch_size}
Batch size is highly related to PLM's training efficiency, and previous work~\cite{liu2019roberta,li2020train,you2019reducing} found that slow-but-accurate large batch sizes can bring improvements to model training, although the improvements become marginal after increasing the batch size beyond a certain point (around $2,048$). BERT~\cite{devlin2018bert} is pre-trained for $1,000$k steps with a batch size of $256$, and the computational cost is equivalent to training for $125$k steps with a batch size of $2,048$~\cite{liu2019roberta}, which is the pre-training setting chosen in our main paper. Choosing $\text{RoBERTa}_{\texttt{MEDIUM}}$ as the teacher model and $\text{RoBERTa}_{\texttt{BASE}}$ as the student model, in Figure \ref{fig:size_time_batch} we compare the validation PPL as we vary the batch size in $\{256, 512, 1024, 2,048\}$, controlling for the number of passes through the pre-training corpus. We also vary the peak learning rate in $\{1.0\times10^{-4}, 2.5\times10^{-4}, 3.8\times10^{-4}, 5.0\times10^{-4}\}$ and pre-train for $\{1,000\text{k}, 500\text{k}, 250\text{k}, 125\text{k}\}$ steps, respectively, when increasing the batch size. We observe that increasing the batch size results in improved final validation PPL, which is aligned with previous findings~\cite{liu2019roberta}. When adjusting batch size, KI accelerates the convergence unanimously, and its benefits become more evident when training with a smaller batch size, reflected in the absolute improvement in final validation PPL. We hypothesize that this is because learning from the smoothed target probability of KI, containing rich \textit{secondary information}~\cite{yang2019training} or \textit{dark knowledge}~\cite{furlanello2018born}, makes the pre-training process more stable. The student PLM is prevented from fitting to unnecessarily strict distributions and can thus learn faster.

\subsection{Additional Experiments of the Effects of Teacher Model $\mathcal{M}_S$'s architecture (width)}
\label{sec:arch_hidden_size}
We show in Figure \ref{fig:gpt_continual} the validation PPL of $\mathcal{M}_L$ when choosing the teacher PLM $\mathcal{M}_S$ with different hidden sizes ($\{384, 480, 576, 672\}$). As mentioned in our main paper, choosing a wider teacher model improves the training efficiency of the student PLM.

\subsection{Additional Experiments of Knowledge Inheritance for Domain Adaptation}

\paragraph{Different Number of Post-training Steps.}

\begin{table}[h]
  \centering
  \small
  \begin{tabular}{c@{~~~~}c@{~~~~}c@{~~~~}c@{~~~~}c@{~~~~}c}
  \toprule
  \textbf{Domain} & \textbf{Strategy} & $3,400$M & $200$M & $100$M & $40$M \\
  \midrule
  \multirow{2}[2]{*}{CS} & \text{SL}  & $6.71$   &  $7.01$   &  $7.39$    &  $8.77$\\
        & \text{KI}    & $8.63$  & $9.39$  & $9.48$  &  $9.87$
        \\
  \midrule
  \multirow{2}[2]{*}{BIO} & \text{SL}     & $7.29$   & $6.61$  & $8.16$  &  $10.34$ \\
        & \text{KI}    & $10.74$  &  $10.78$  &  $10.93$  & $11.66$ \\
  \bottomrule
  \end{tabular}%
  \caption{The validation PPL on the source domain (WB) after $\text{RoBERTa}_{\texttt{BASE\_WB}}$ is post-trained on the target domain (CS / BIO) with self-learning (SL) and knowledge inheritance (KI).}
  \label{tab:continual_source}%
\end{table}%

\begin{figure*}[!t]
    \centering
    \subfigure{\includegraphics[width=0.3\textwidth]{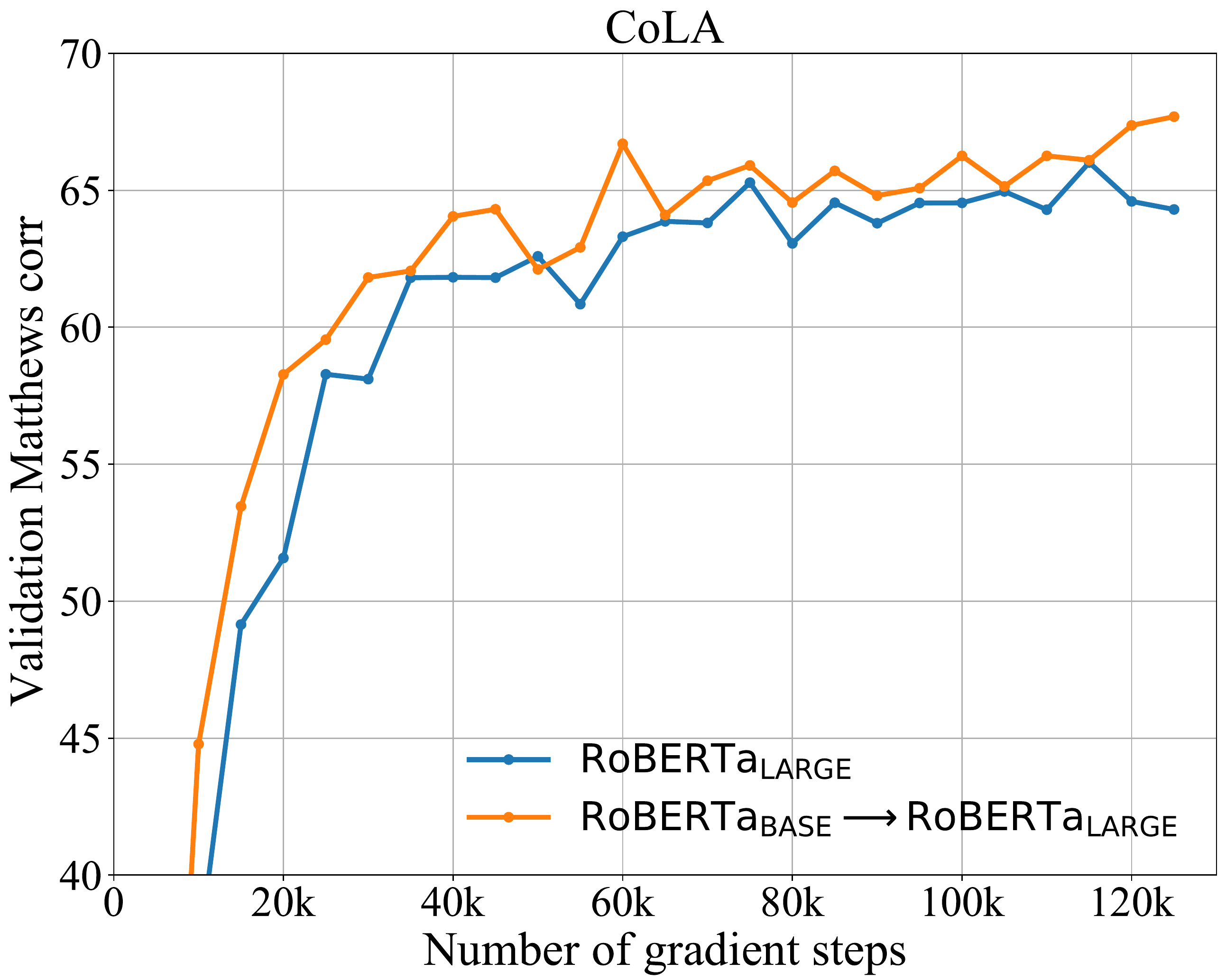}} 
    \subfigure{\includegraphics[width=0.3\textwidth]{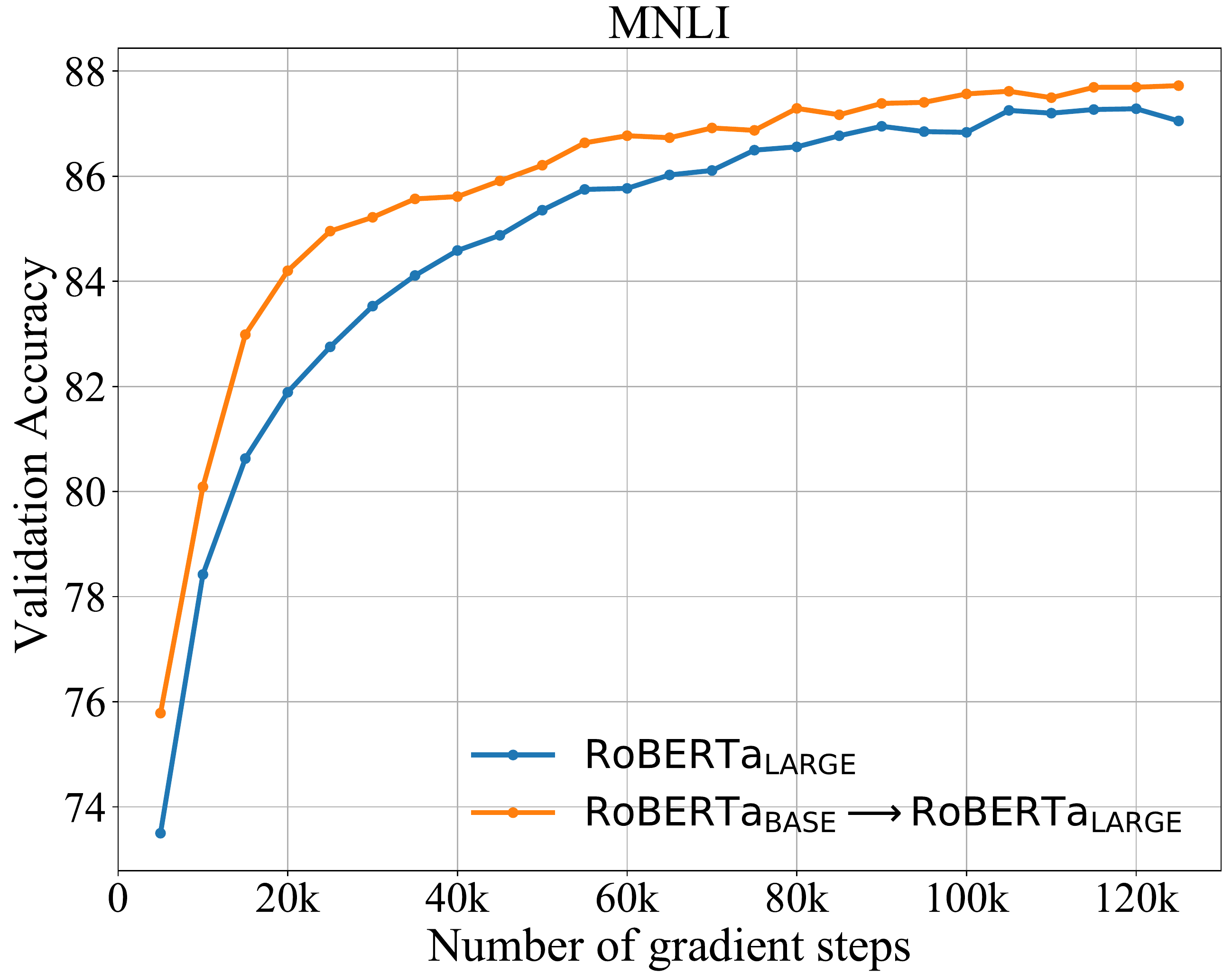}} 
    \subfigure{\includegraphics[width=0.3\textwidth]{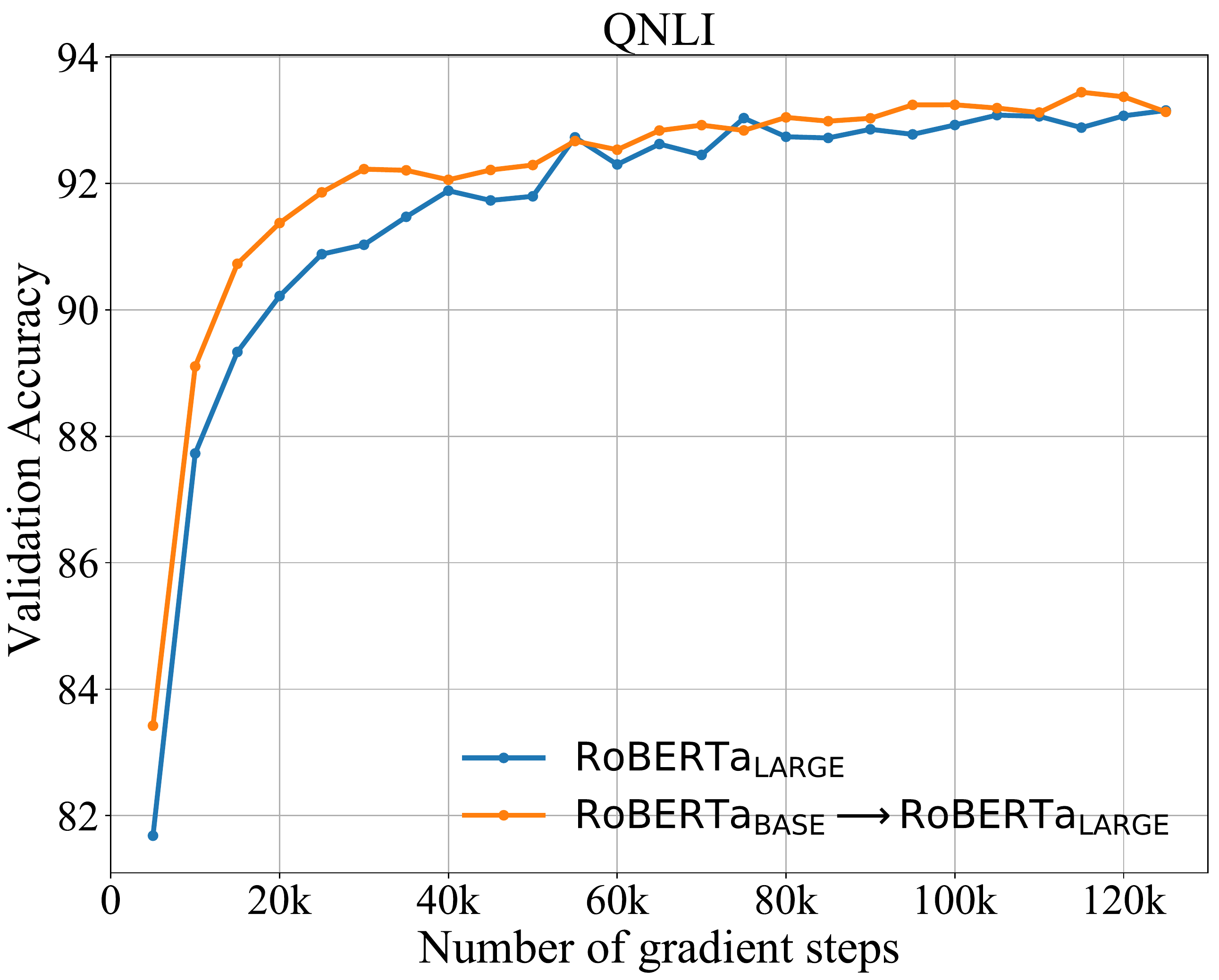}}
    \subfigure{\includegraphics[width=0.3\textwidth]{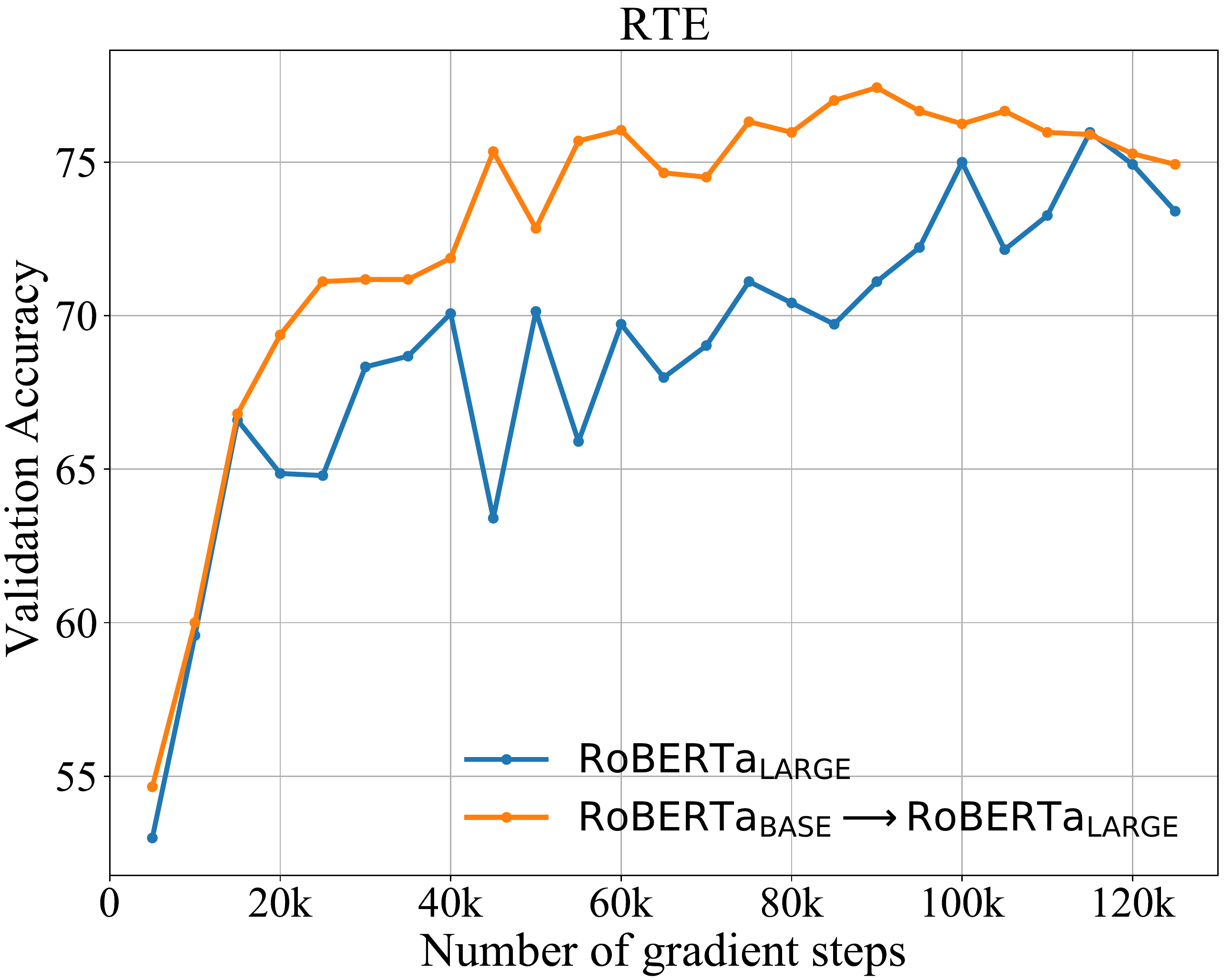}}
    \subfigure{\includegraphics[width=0.3\textwidth]{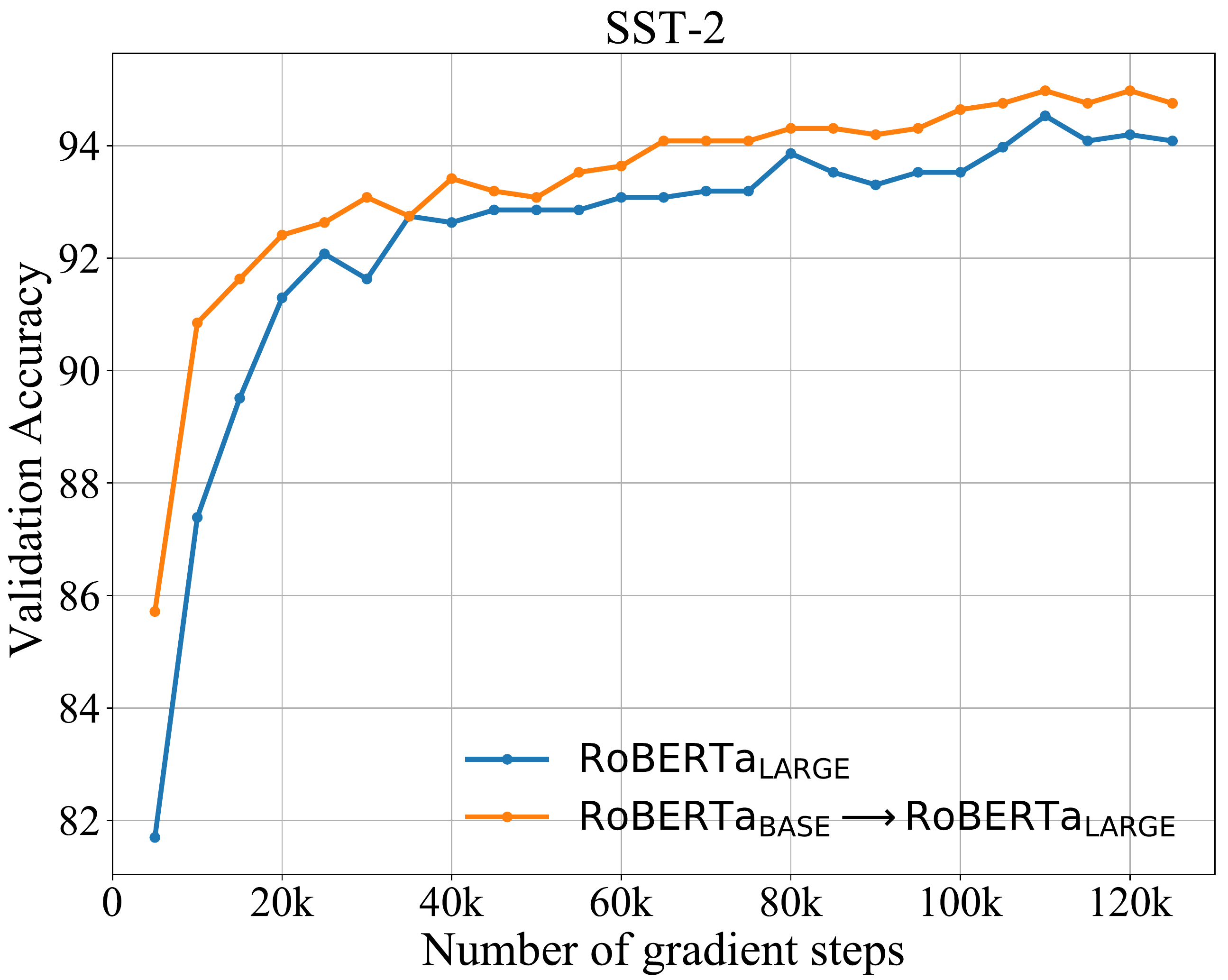}}
    \subfigure{\includegraphics[width=0.3\textwidth]{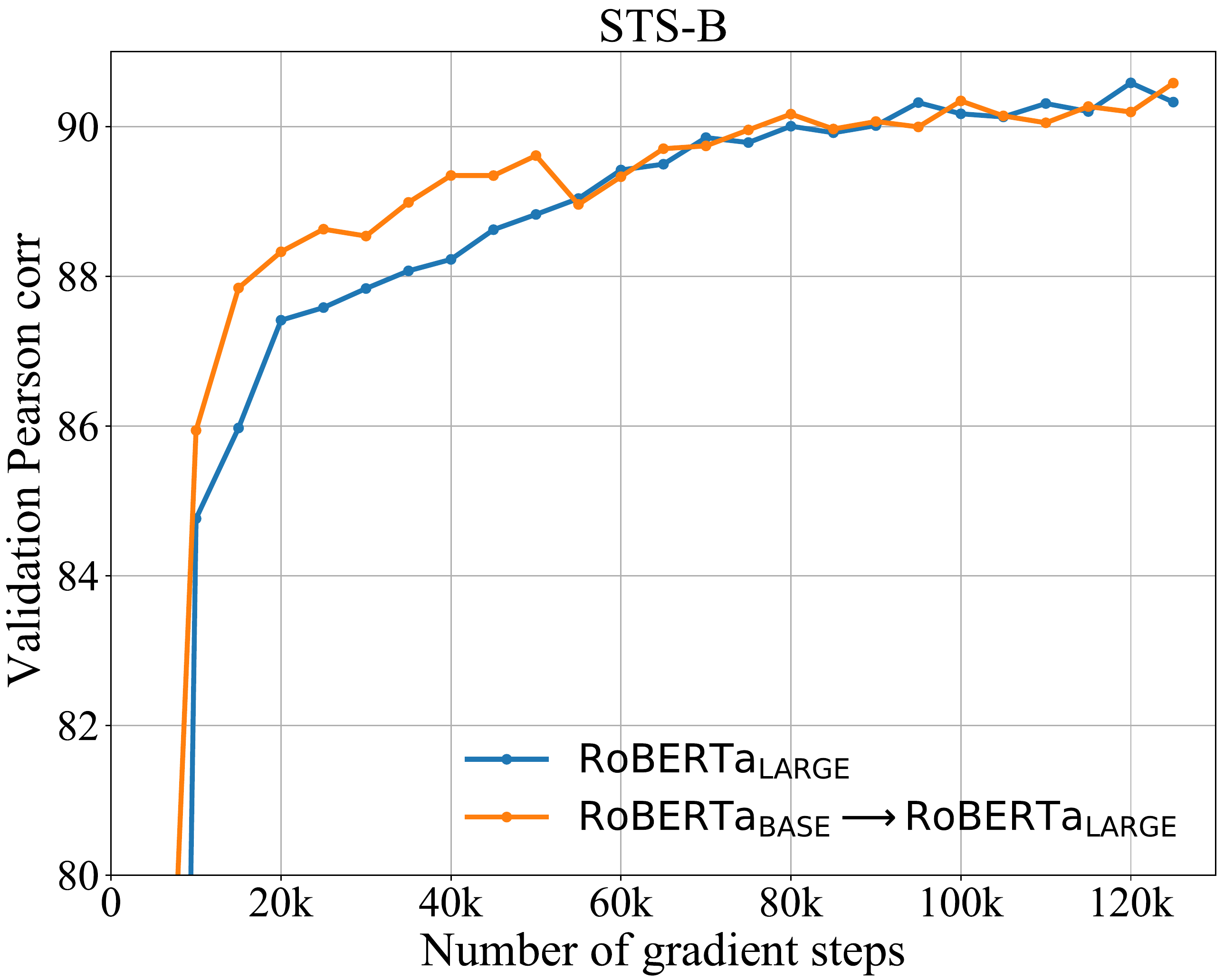}}
    \caption{Downstream performance visualization on six GLUE tasks comparing $\text{RoBERTa}_{\texttt{LARGE}}$ and $\text{RoBERTa}_{\texttt{BASE}} \rightarrow \text{RoBERTa}_{\texttt{LARGE}}$. For CoLA, RTE, SST-2 and STS-B, we repeat fine-tuning for $5$ times; for MNLI and QNLI, we repeat fine-tuning for $3$ times.}
    \label{fig:fine-tune}
\end{figure*}

In the main paper, we adapt $\text{RoBERTa}_{\texttt{BASE\_WB}}$ to either CS or BIO domain by post-training it for $4$k steps. We further vary the number of training steps in $\{1\text{k}, 2\text{k}, 3\text{k}, 4\text{k}, 5\text{k}\}$ and visualize the validation PPL in Figure \ref{fig:gpt_continual}. We also experiment on different sizes of domain corpus, i.e., $3,400$M, $200$M, $100$M, $40$M tokens, respectively, as done in the main paper. We observe that generally the validation PPL on each domain decreases with the training step growing, and the performance of KI is always better than self-learning. The improvement of KI over self-learning is further enlarged when there is less target domain data available, demonstrating that KI is more data-efficient and can work well in low-resource settings. In addition, self-learning exhibits overfitting problems when the data size of the target domain is relatively small, which is not observed under our KI framework, which means KI can mitigate overfitting under low-resource settings.


\paragraph{Catastrophic Forgetting on the Source Domain.}
\label{sec:catastrophic}
Table \ref{tab:continual_source} lists the validation PPL on the source domain (WB) after $\text{RoBERTa}_{\texttt{BASE\_WB}}$ is post-trained on the target domain (CS / BIO) with self-learning (SL) and knowledge inheritance (KI) for $4$k steps. We show the results w.r.t. different sizes of domain corpus ($3,400$M, $200$M, $100$M and $40$M tokens). We observe that after domain adaptation, the validation PPL on the source domain increases, which means PLMs may forget some key knowledge on the source domain when learning new knowledge in the target domain, i.e., the catastrophic forgetting problem. In addition, we find that the problem is more evident for KI than self-learning. We expect future work to further explore how to mitigate the catastrophic forgetting.



\subsection{Detailed Downstream Performances on GLUE Tasks}
\label{sec:downstream_detail}

\begin{figure*}
    \centering
    \subfigure{\includegraphics[width=0.3\textwidth]{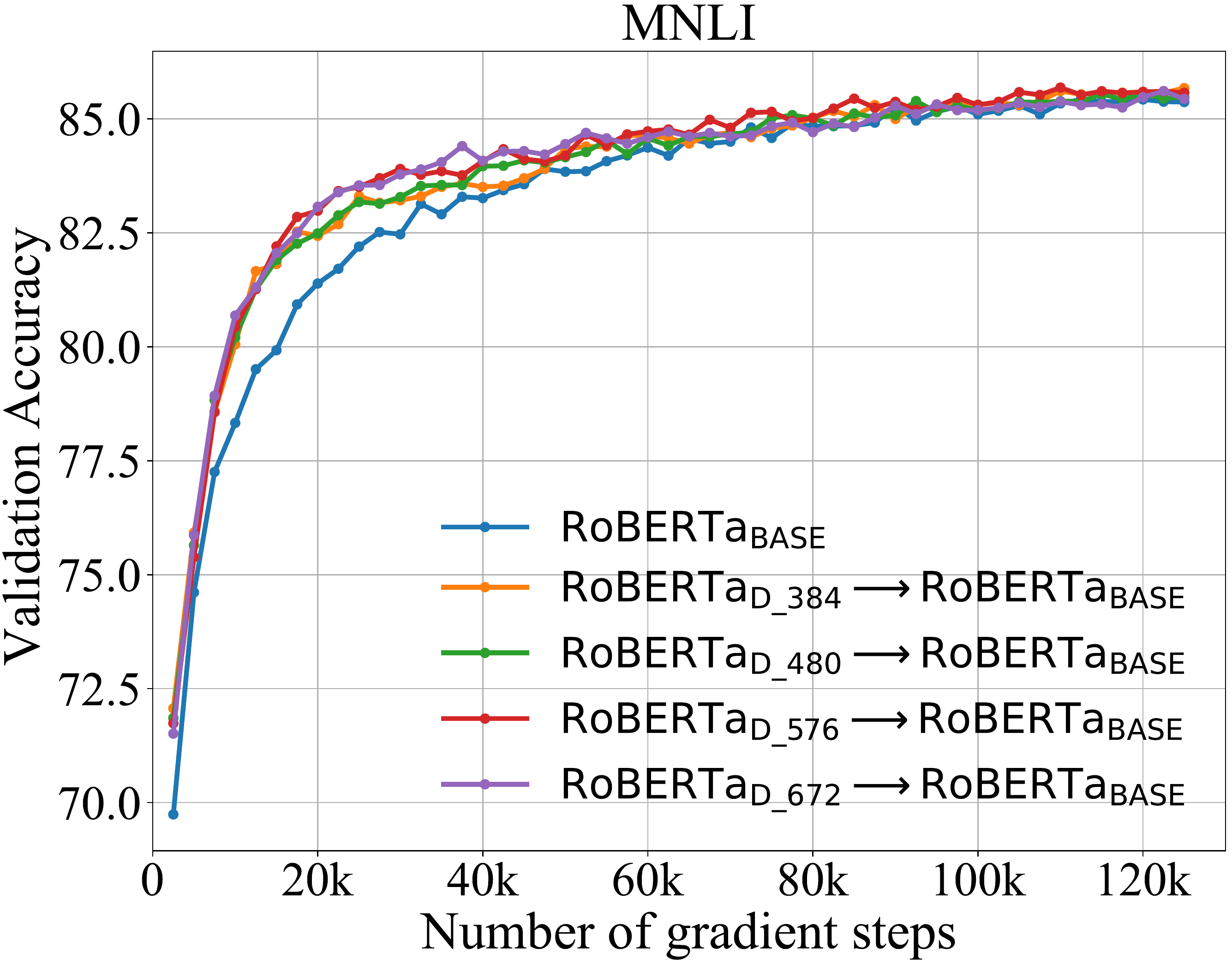}} 
    \subfigure{\includegraphics[width=0.3\textwidth]{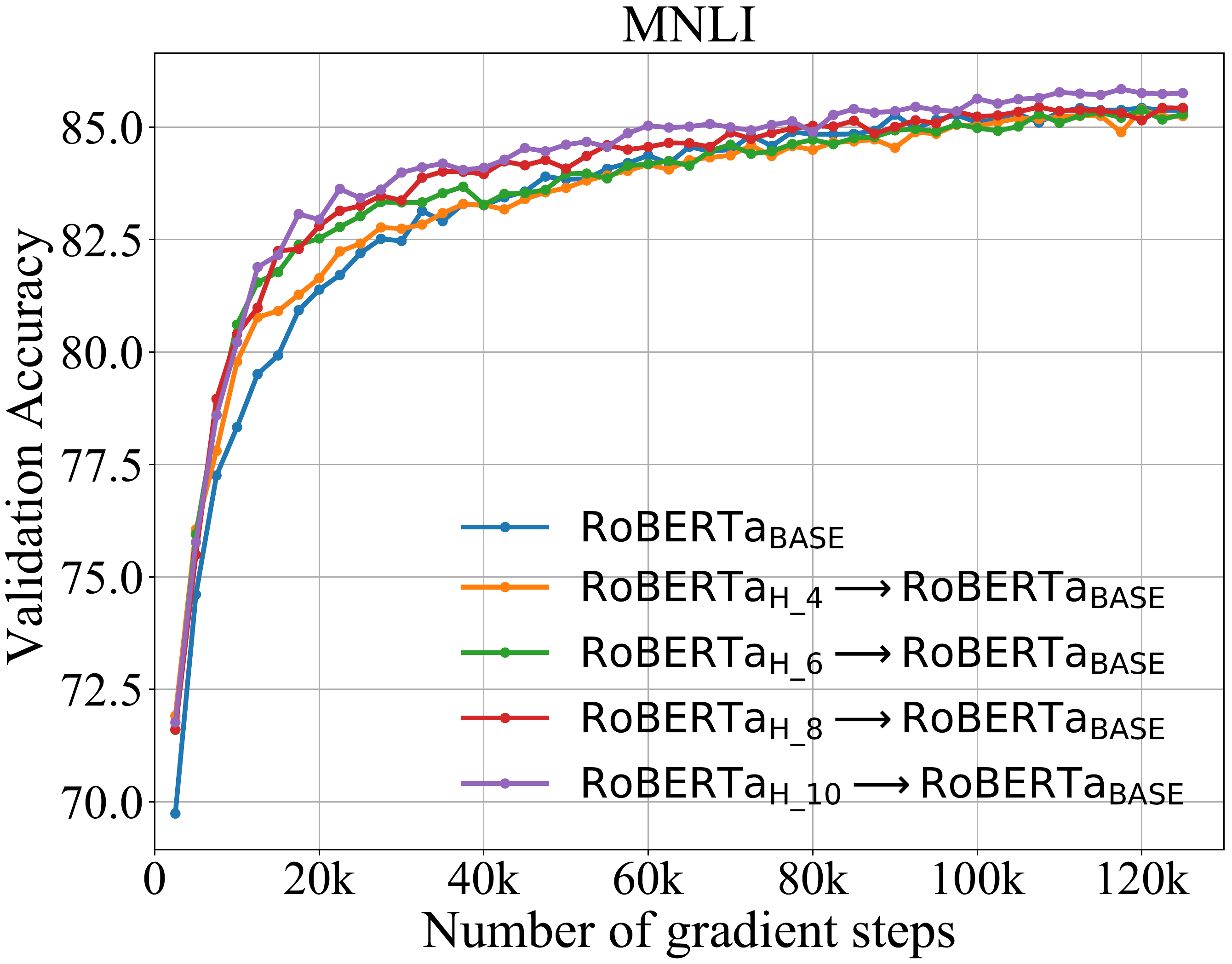}} 
    \subfigure{\includegraphics[width=0.3\textwidth]{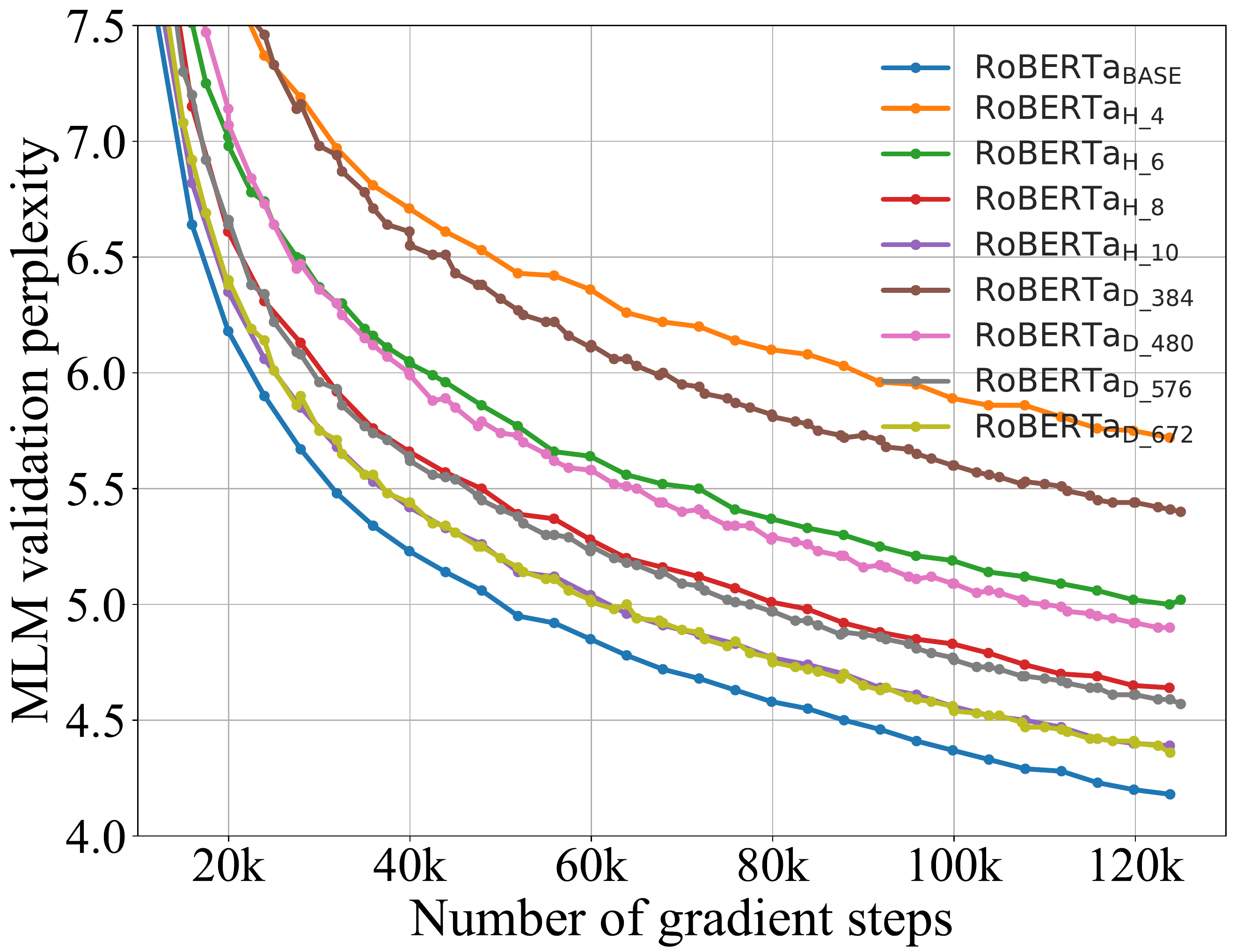}} 
    \caption{Left \& Middle: downstream performances corresponding to the experiments on effects of $\mathcal{M}_S$'s model architecture (width (left) \& depth (middle)). Right: validation PPL during pre-training for the teacher models used in experiments of effects of teacher model architecture.}
    \label{fig:mnli_arch}
\end{figure*}
Figure \ref{fig:fine-tune} visualizes the downstream performance of $\text{RoBERTa}_{\texttt{LARGE}}$ and $\text{RoBERTa}_{\texttt{BASE}} \rightarrow \text{RoBERTa}_{\texttt{LARGE}}$ on the dev sets of six GLUE tasks at different pre-training steps with an interval of $5$k. It can be observed that the downstream performance of $\text{RoBERTa}_{\texttt{BASE}} \rightarrow \text{RoBERTa}_{\texttt{LARGE}}$ rises faster than the baseline, which means it takes fewer pre-training steps for our KI framework to get a high score in downstream tasks. Aligned with previous findings~\cite{li2020train}, we found MNLI and SST-2 to be the most stable tasks in GLUE, whose variances are lower.

We also list the average GLUE performance for $\text{RoBERTa}_{\texttt{BASE}} \rightarrow \text{RoBERTa}_{\texttt{LARGE}}$ and the baseline $\text{RoBERTa}_{\texttt{LARGE}}$ in Table \ref{tab:full_main_glue}, from which we observe that the baseline at $70$k-th step achieves almost the same GLUE performance as our method at $40$k-th step, which means our framework saves around $42.9$\% FLOPs, much higher than the reported $27.3$\% FLOPs saved based on the pre-training PPL metric in the main paper. In addition, our method achieves almost the same GLUE performance as the baseline at the final step ($125$k) with only $70$k steps, which means our framework saves $44$\% FLOPs in total. Both the perplexity in the pre-training stage and performance in downstream tasks can be chosen as the evaluation metric for measuring the computational cost savings. However, in this paper, we choose the former because it is more stable and accurate than the latter. We find empirically that some GLUE tasks like CoLA have higher variances than others, which might make the measurement inaccurate.

Besides, when discussing the effects of model architectures in the main paper, we only show the validation PPL of each model during pre-training, we visualize the corresponding downstream performance (MNLI) in Figure \ref{fig:mnli_arch}, from which it can be observed that learning from teacher models with more parameters helps achieve better downstream performance at the same pre-training step. In general, we observe that, under our setting, the performance gain in downstream tasks is aligned with that reflected in validation PPL during pre-training.  

\begin{table}[!tbp]
  \centering
  \small
    \begin{tabular}{c@{~~}c@{~~}c}
    \toprule
    \textbf{Step}   & $\text{RoBERTa}_{\texttt{BASE}}$  & $\text{RoBERTa}_{\texttt{BASE}} \rightarrow \text{RoBERTa}_{\texttt{LARGE}}$  \\
    \midrule
    $5$k   & $61.8$  & $68.8$  \\
    $10$k   & $75.6$  & $78.1$  \\
    $15$k   & $79.3$  & $81.5$  \\
    $20$k   & $80.4$  & $82.8$  \\
    $25$k   & $81.7$  & $83.6$  \\
    $30$k   & $82.4$  & $83.9$  \\
    $35$k   & $83.1$  & $84.1$  \\
    $40$k   & $83.6$  & $84.5$  \\
    $45$k   & $82.8$  & $85.2$  \\
    $50$k   & $83.9$  & $84.6$  \\
    $55$k   & $83.4$  & $85.2$  \\
    $60$k   & $84.0$  & $85.7$  \\
    $65$k   & $84.1$  & $85.3$  \\
    $70$k   & $84.3$  & $85.5$  \\
    $75$k   & $85.0$  & $85.8$  \\
    $80$k   & $84.7$  & $85.8$  \\
    $85$k   & $84.8$  & $86.2$  \\
    $...$   & $...$  & $...$  \\
    $125$k   & $85.5$  & $86.1$  \\
    \bottomrule
    \end{tabular}%
  \caption{Average GLUE performance comparing both $\text{RoBERTa}_{\texttt{BASE}}$ and $\text{RoBERTa}_{\texttt{BASE}} \rightarrow \text{RoBERTa}_{\texttt{LARGE}}$ at different pre-training steps.}
  \label{tab:full_main_glue}%
\end{table}%

\subsection{Teacher Models' Validation PPL Curves during Pre-training for ``Effects of Model Architecture''}
\label{sec:exp_teacher_ppl}
Figure \ref{fig:mnli_arch} visualizes the validation PPL curves for all the teacher models used in the experiments on the effects of model architecture. The teacher models differ from $\text{RoBERTa}_{\texttt{BASE}}$ in either the depth or width. Specifically, we vary the depth in $\{4,6,8,10\}$ (denoted as $\{\text{RoBERTa}_{\texttt{H\_4}}$, $\text{RoBERTa}_{\texttt{H\_6}}$, $\text{RoBERTa}_{\texttt{H\_8}}$, $\text{RoBERTa}_{\texttt{H\_10}}\}$), and the width in $\{384, 480, 576, 672\}$ (denoted as $\{\text{RoBERTa}_{\texttt{D\_384}}$, $\text{RoBERTa}_{\texttt{D\_480}}$, $\text{RoBERTa}_{\texttt{D\_576}}$, $\text{RoBERTa}_{\texttt{D\_672}}\}$). Generally, PLMs with larger model parameters converge faster and achieve better final performance.

\begin{table*}[h]
  \centering
  \small
    \begin{tabular}{lccccccc}
    \toprule
            \multicolumn{1}{l}{$\textbf{Model Name}$}
          & \multicolumn{1}{l}{$n_{\text{params}}$} & \multicolumn{1}{l}{$n_{\text{layers}}$} & \multicolumn{1}{l}{$d_{\text{model}}$} & \multicolumn{1}{l}{$n_{\text{heads}}$} & \multicolumn{1}{l}{$d_{\text{FFN}}$} & \multicolumn{1}{l}{\text{lr ($\text{bs} = 2,048$)}} \\
    \midrule
    $\text{RoBERTa}_{\texttt{MEDIUM}}$ & $74$M      & $9$      & $576$      &  $12$     &   $3072$    &  $5.0\times10^{-4}$  \\
    $\text{RoBERTa}_{\texttt{D\_d}}$ & -      & $12$      & $\texttt{d}$      &  $12$     &   $3072$    &  $5.0\times10^{-4}$  \\
    $\text{RoBERTa}_{\texttt{H\_h}}$ & -      & $\texttt{h}$      & $768$      &  $12$     &   $3072$    &  $5.0\times10^{-4}$  \\
    $\text{RoBERTa}_{\texttt{BASE}}$ &  $125$M     & $12$   &  $768$ &  $12$ &  $3072$  &  $5.0\times10^{-4}$ \\
    $\text{RoBERTa}_{\texttt{BASE\_PLUS}}$ &  $211$M     & $18$   &  $864$ &  $12$ &  $3600$  &  $3.5\times10^{-4}$  \\
    $\text{RoBERTa}_{\texttt{LARGE}}$ &  $355$M     & $24$   &  $1024$ &  $16$ &  $4096$  &  $2.5\times10^{-4}$  \\
    \midrule
    $\text{GPT}_{\texttt{73M}}$ &   $73$M    &  $9$     &  $576$     & $12$      & $3072$      &  $5.0\times10^{-4}$  \\
    $\text{GPT}_{\texttt{124M}}$ &   $124$M    &  $12$     &  $768$     & $12$      & $3072$      &  $5.0\times10^{-4}$ \\
    $\text{GPT}_{\texttt{209M}}$ &  $209$M     &  $18$     &  $864$     & $12$      & $3600$      &  $4.0\times10^{-4}$  \\
    $\text{GPT}_{\texttt{354M}}$ &  $354$M     &  $24$     &  $1024$     & $16$      & $4096$      &  $3.5\times10^{-4}$  \\
    $\text{GPT}_{\texttt{773M}}$ &  $773$M     &  $36$     &  $1280$     & $20$      & $5120$      &  $3.0\times10^{-4}$  \\
    $\text{GPT}_{\texttt{1B}}$ &  $1068$M     &  $40$     &  $1440$     & $20$      & $5760$      &  $2.5\times10^{-4}$  \\
    \bottomrule
    \end{tabular}%
  \caption{Model architectures for all the models we used in this paper.}
  \label{tab:arch}%
\end{table*}%

\begin{table*}[h]
  \centering
  \small
    \begin{tabular}{ccc}
    \toprule
    $\mathcal{M}_L$ & $\mathcal{M}_S$ & Steps of teacher-guided learning \\
    \midrule
    \multirow{9}[2]{*}{$\text{RoBERTa}_{\texttt{BASE}}$} & $\text{RoBERTa}_{\texttt{MEDIUM}}$ &  $35$k \\
          & $\text{RoBERTa}_{\texttt{D\_384}}$ & $28$k \\
          & $\text{RoBERTa}_{\texttt{D\_480}}$ & $40$k \\
          & $\text{RoBERTa}_{\texttt{D\_576}}$ & $70$k \\
          & $\text{RoBERTa}_{\texttt{D\_672}}$ & $85$k \\
          & $\text{RoBERTa}_{\texttt{H\_4}}$  &  $22$k \\
          & $\text{RoBERTa}_{\texttt{H\_6}}$  &  $35$k \\
          & $\text{RoBERTa}_{\texttt{H\_8}}$  &  $55$k \\
          & $\text{RoBERTa}_{\texttt{H\_10}}$ &  $65$k \\
    \midrule
    $\text{RoBERTa}_{\texttt{BASE\_PLUS}}$ & $\text{RoBERTa}_{\texttt{BASE}}$  &  $55$k \\
    \midrule
    \multirow{3}[2]{*}{$\text{RoBERTa}_{\texttt{LARGE}}$} & $\text{RoBERTa}_{\texttt{BASE}}$  & $40$k \\
          & $\text{RoBERTa}_{\texttt{BASE\_PLUS}}$ & $65$k \\
          & $\text{RoBERTa}_{\texttt{BASE}} \rightarrow \text{RoBERTa}_{\texttt{BASE\_PLUS}}$ & $75$k \\
    \midrule
    $\text{GPT}_{\texttt{124M}}$  & $\text{GPT}_{\texttt{73M}}$ &  $10$k \\
    \midrule
    $\text{GPT}_{\texttt{209M}}$ & $\text{GPT}_{\texttt{124M}}$  &  $15$k \\
    \midrule
    $\text{GPT}_{\texttt{354M}}$ & $\text{GPT}_{\texttt{209M}}$  &  $18$k \\
    \midrule
    $\text{GPT}_{\texttt{773M}}$ & $\text{GPT}_{\texttt{354M}}$  &  $16$k \\
    \midrule
    $\text{GPT}_{\texttt{1B}}$ & $\text{GPT}_{\texttt{773M}}$  &  $20$k \\
    \bottomrule
    \end{tabular}%
  \caption{The total number of steps for teacher-guided learning for different ($\mathcal{M}_L$, $\mathcal{M}_S$) pairs.}
  \label{tab:teaching_phase}%
\end{table*}%

\section{Pre-training Hyper-parameters}
\label{sec:hyper_pretrain}
In Table \ref{tab:arch}, we list the architectures we used for all models, covering the details for the total number of trainable parameters ($n_{\text{params}}$), the total number of layers ($n_{\text{layers}}$), the number of units in each bottleneck layer ($d_{\text{model}}$), the total number of attention heads ($n_{\text{heads}}$), the inner hidden size of FFN layer ($d_{\text{FFN}}$) and the learning rate when batch size is set to $2,048$ (lr). The training-validation ratio of pre-training data is set to $199:1$. We set the weight decay to $0.01$, dropout rate to $0.1$, and use linear learning rate decay. Adam is chosen as the optimizer. The learning rate is warmed up for the first $10$\% steps. The hyper-parameters for Adam optimizer is set to $1\times10^{-6}, 0.9, 0.98$ for $\epsilon, \beta_1, \beta_2$, respectively. For a fair comparison, all experiments are done in the same computation environment with $8$ NVIDIA 32GB V100 GPUs. Table \ref{tab:teaching_phase} describes the total number of pre-training steps for each ($\mathcal{M}_L$, $\mathcal{M}_S$) pair chosen in our experiments.

\section{Fine-tuning Hyper-parameters}

\begin{table*}[thbp]
  \centering
  \small
    \begin{tabular}{l@{~~~}c@{~~~}c@{~~~}}
    \toprule
    \textbf{HyperParam} & ACL-ARC \& CHEMPROT & GLUE \\
    \midrule
    Learning Rate &  $2\times10^{-5}$     & $\{1\times10^{-5}, 2\times10^{-5}, 3\times10^{-5}\}$ \\
    Batch Size &   $256$    &   $\{16, 32\}$ \\
    Weight Decay &  $0.1$     &  $0.1$ \\
    Max Epochs &    $10$   &   $10$ \\
    Learning Rate Decay &  Linear     &  Linear \\
    Warmup Ratio &  $0.06$     &   $0.06$ \\
    \bottomrule
    \end{tabular}%
  \caption{Hyper-parameters for fine-tuning $\text{RoBERTa}$ on ACL-ARC, CHEMPROT and GLUE.}
  \label{tab:finetune}%
\end{table*}%

Table \ref{tab:finetune} describes the hyper-parameters for ACL-ARC, CHEMPROT and GLUE tasks. The selection of these hyper-parameters closely follows \cite{liu2019roberta} and \cite{gururangan2020don}. 

\section{Domain Proximity of WB, CS and BIO}
\label{sec:domain_proximity}
\begin{table}[thbp]
  \centering
  \small
    \begin{tabular}{l|ccc}
    \toprule
          & WB & CS & BIO \\
    \midrule
    WB    & $100$\%   & $19.1$\%  & $25.6$\% \\
    CS    & $19.1$\%  & $100$\%   & $22.5$\% \\
    BIO   & $25.6$\%  & $22.5$\%  & $100$\% \\
    \bottomrule
    \end{tabular}%
  \caption{Domain proximity (vocabulary overlap) among three domains (WB, CS, BIO) discussed in this paper. Following \cite{gururangan2020don}, we create the vocabulary for each domain by considering the top $10$k most frequent words (excluding stopwords).}
  \label{tab:proximity}%
\end{table}%

Table \ref{tab:proximity} lists the domain proximity (vocabulary overlap) of WB, CS and BIO used in this paper.



\section{Comparison between Knowledge Inheritance and Parameter Recycling}
\label{sec:compare_KI_PT}
Parameter recycling (i.e., progressive training) first trains a small PLM, and then gradually increases the depth or width of the network based on parameter initialization. It is an orthogonal research direction against our KI, and has many limitations as follows:

\paragraph{Architecture Mismatch.} Existing parameter recycling methods~\citep{gong2019efficient,gu2020transformer} require that the architectures of both small PLMs and large PLMs are matched to some extent, however, our KI does not have such a requirement. For example, \citet{gong2019efficient,gu2020transformer} either requires the number of layers, or the hidden size/embedding size of a large PLM to be the integer multiples of that of a small PLM. Hence, it is not flexible to train larger PLMs with arbitrary architectures, making parameter recycling hard to be implemented practically. Besides, there are more and more advanced non-trivial Transformer modifications appearing (we refer to \citet{lin2021survey} for details), e.g., pre-normalization, relative embedding, sparse attention, etc. It is non-trivial to directly transfer the parameters between two PLMs if they have different inner structures. Nevertheless, our KI framework will not be influenced by such architectural mismatches.

\paragraph{Inability for Multi-to-one Knowledge Inheritance.} It is non-trivial to support absorbing knowledge from multiple teacher models by jointly recycling their model parameters. Instead, it is easy to implement for KI. As shown in our experiments, we demonstrate that under our framework, large PLMs can simultaneously absorb knowledge from multiple teachers.

\paragraph{Inability of Knowledge Inheritance for Domain Adaptation.} Parameter recycling is hard to support continual learning, which makes large PLMs absorb knowledge from small ones in a lifelong manner. In real-world scenarios, numerous PLMs of different architectures are trained locally with different data. These small PLMs can be seen as domain experts, and it is essential that larger PLMs can continuously benefit from these existing PLMs efficiently by incorporating their knowledge so that larger PLMs can become omnipotent. As described before, it is easy to implement for our framework and we have demonstrated the effectiveness.

\paragraph{Model Privacy.} Parameter recycling requires the availability of the parameters of an existing PLM, which may be impractical due to some privacy issues, e.g., GPT-3 only provides API access for prediction instead of the model parameters. Instead, our KI framework does not presume access to an existing model parameter since the predictions of the small model can be pre-computed and saved offline. This superiority will further make it possible for API-based online knowledge transfer.

\section{Comparing Label Smoothing and Knowledge Inheritance}
\label{sec:exp_label_smoothing}

Previous work shows the relation between label smoothing and knowledge distillation to some extent~\cite{shen2021label}. To demonstrate that the success of our KI is not because of learning from a more smoothed target, we conduct experiments comparing both label smoothing and our KI in Table \ref{tab:label_smoothing}. Specifically, for label smoothing, PLMs optimize a smoothed target $\textbf{y}_i^S = (1 - \alpha) * \textbf{y}_i + \alpha * \Vec{\mathbf{1}} / (K-1)$, where $\alpha = 0$ denotes learning from scratch with no label smoothing, larger $\alpha$ means a more smoothed target for PLMs to learn from, $K$ denotes the vocabulary size. Specifically, we choose $\alpha$ from $\{0.1, 0.2, 0.3\}$. It can be concluded from the results in Table~\ref{tab:label_smoothing} that adding label smoothing into the pre-training objectives of PLMs leads to far worse performance than the vanilla baseline, which shows that the improvements of our knowledge inheritance framework are non-trivial: larger PLMs are indeed inheriting the ``knowledge'' from smaller ones, instead of benefiting from optimizing a smoothed target, which imposes regularization.

\begin{table}[!tbp]
  \centering
  \small
    \begin{tabular}{cccccc}
    \toprule
    \textbf{Step}  & \multicolumn{1}{c}{$20$k} & \multicolumn{1}{c}{$40$k} & \multicolumn{1}{c}{$60$k} & \multicolumn{1}{c}{$80$k} & \multicolumn{1}{c}{$100$k} \\
    \midrule
    $\alpha = 0.3$ & $8.68$  & $7.29$  & $6.90$   & $6.57$  & $6.26$ \\
    $\alpha=0.2$ & $7.27$  & $6.47$  & $5.95$  & $5.68$  & $5.46$ \\
    $\alpha=0.1$ & $6.71$  & $5.74$  & $5.35$  & $5.06$  & $4.86$ \\
    $\alpha=0$ & $6.13$  & $5.21$  & $4.83$  & $4.57$  & $4.36$ \\
    $\textbf{KI}$    & $\mathbf{5.69}$  & $\mathbf{5.17}$  & $\mathbf{4.78}$  & $\mathbf{4.52}$  & $\mathbf{4.32}$ \\
    \bottomrule
    \end{tabular}%
  \caption{Validation PPL for training $\text{RoBERTa}_{\texttt{BASE}}$ with different strategies. $\mathbf{KI}$ denotes our knowledge inheritance framework, where $\text{RoBERTa}_{\texttt{MEDIUM}}$ is chosen as the teacher.}
  \label{tab:label_smoothing}%
\end{table}%


\end{document}